\documentclass[10pt,twocolumn,letterpaper]{article}

\usepackage{iccv}
\usepackage{times}
\usepackage{epsfig}
\usepackage{graphicx}
\usepackage{amsmath}
\usepackage{amssymb}
\usepackage{appendix}

\usepackage[pagebackref=true,breaklinks=true,letterpaper=true,colorlinks,bookmarks=false]{hyperref}

\iccvfinalcopy 

\def\iccvPaperID{8900} 

\ificcvfinal\fi

\begin{document}

\title{AA-RMVSNet: Adaptive Aggregation Recurrent Multi-view Stereo Network}

\author{Zizhuang Wei, Qingtian Zhu, Chen Min, Yisong Chen and Guoping Wang\thanks{Corresponding author.}\\
Peking University\\
{\tt\small \{weizizhuang,wgp\}@pku.edu.cn}
}

\maketitle
\ificcvfinal\thispagestyle{empty}\fi

\begin{abstract}
In this paper, we present a novel recurrent multi-view stereo network based on long short-term memory (LSTM) with adaptive aggregation, namely AA-RMVSNet.
We firstly introduce an intra-view aggregation module to adaptively extract image features by using context-aware convolution and multi-scale aggregation, which efficiently improves the performance on challenging regions, such as thin objects and large low-textured surfaces.
To overcome the difficulty of varying occlusion in complex scenes, we propose an inter-view cost volume aggregation module for adaptive pixel-wise view aggregation, which is able to preserve better-matched pairs among all views.
The two proposed adaptive aggregation modules are lightweight, effective and complementary regarding improving the accuracy and completeness of 3D reconstruction.
Instead of conventional 3D CNNs, we utilize a hybrid network with recurrent structure for cost volume regularization, which allows high-resolution reconstruction and finer hypothetical plane sweep.
The proposed network is trained end-to-end and achieves excellent performance on various datasets.
It ranks $1^{st}$ among all submissions on Tanks and Temples benchmark and achieves competitive results on DTU dataset, which exhibits strong generalizability and robustness.
Implementation of our method is available at \url{https://github.com/QT-Zhu/AA-RMVSNet}.
\end{abstract}

\section{Introduction}

\begin{figure}[t]
    \centering
    \includegraphics[width=\columnwidth]{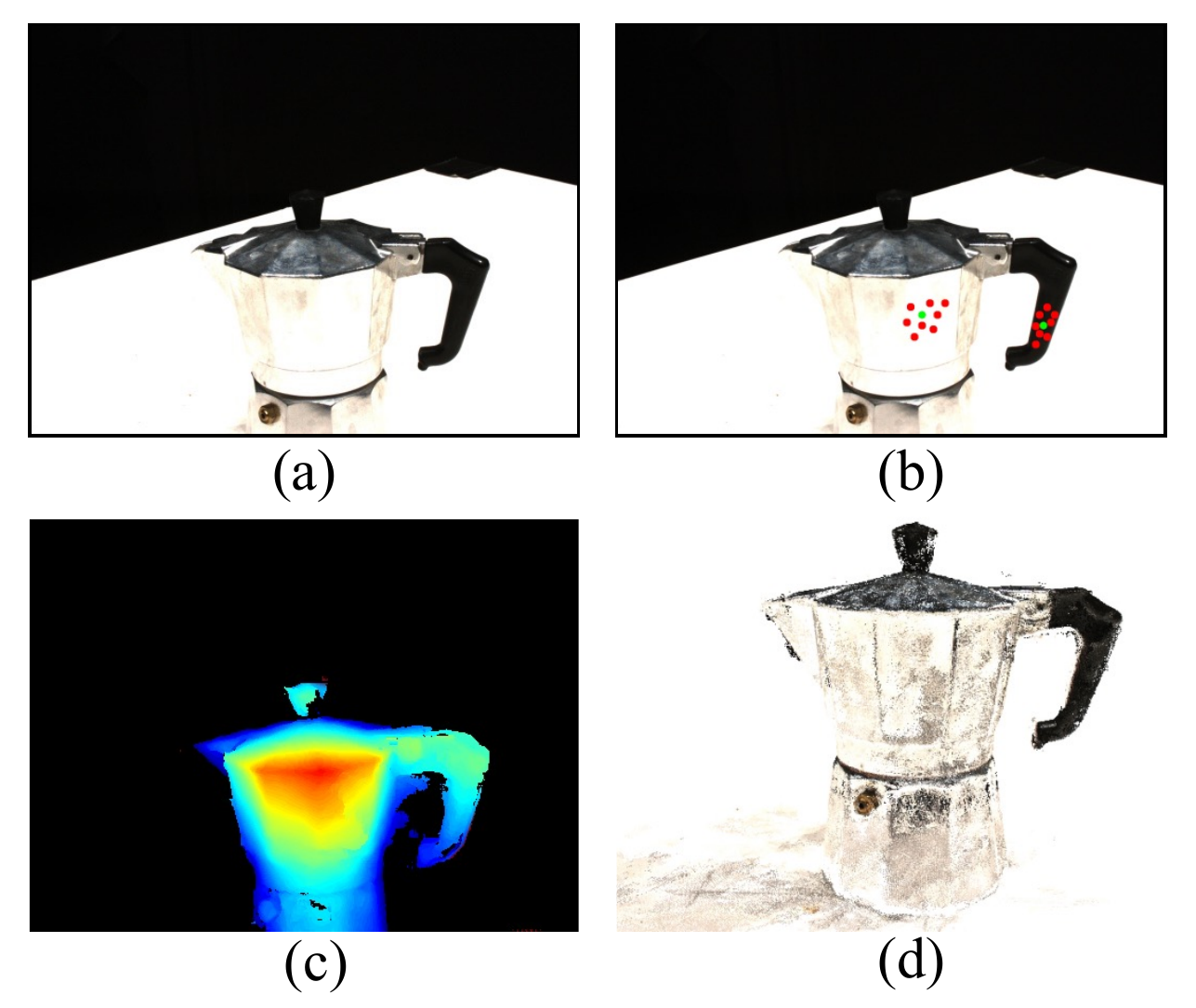}
    \caption{Illustration of multi-view 3D reconstruction of Scan 77 in DTU dataset~\cite{Aans2016Large} using the proposed AA-RMVSNet. (a) The reference image; (b) adaptive sampling locations in our intra-view AA approach; (c) the depth map estimated by AA-RMVSNet after filtering; (d) the recovered dense 3D model.}
    \label{first}
\end{figure}

Multi-view stereo (MVS) aims to obtain 3D dense models of real-world scenes from multiple images, which is one of the core techniques in a variety of applications including virtual reality, autonomous driving and heritage conservation.
While traditional MVS methods~\cite{campbell2008using,furukawa2009accurate,galliani2015massively,schonberger2016mvs,barnes2009patchmatch} utilize hand-crafted matching metrics to measure multi-view consistency, recent deep-learning-based methods~\cite{yao2018mvsnet,yao2019recurrent,2020attention,zhang2020visibility,yan2020dense} achieve superior accuracy and completeness on many MVS benchmarks~\cite{Aans2016Large,knapitsch2017tanks,yao2020blendedmvs,schops2017multi} compared with the previous state-of-the-arts, through introducing convolutional neural network (CNN) which makes feature extraction and cost volume regularization more powerful.
However, some challenging problems still remain to be solved to further improve reconstruction quality. 

First, general features extracted by 2D CNN in regular pixel grids with fixed receptive fields often have difficulties in handling thin structures or textureless surfaces, which limits the robustness and completeness of 3D reconstruction.
Recent MVSNet-based attempts~\cite{yi2020PVAMVSNET,yan2020dense,yang2020cost} introduce multi-scale information to improve depth estimation.
However, context-aware features have not been leveraged well enough for varying richness of texture on different regions.

Second, few works consider pixel-wise visibility issues during multi-view matching cost aggregation, which inevitably deteriorates the final reconstruction quality, especially under severe occlusion.
In order to select well-captured views for each pixel, Vis-MVSNet~\cite{zhang2020visibility}
uses pair-wise matching uncertainties as weighting guidance to attenuate pixels that have difficulties to match.
PVA-MVSNet~\cite{yi2020PVAMVSNET} contains a CNN-based voxel-wise view aggregation module to guide multiple cost volume aggregation.
However, it is hard to give a perfect solution for the occlusion problem in general case.

Moreover, in order to meet the needs of various real-world applications, memory consumption is also essential for a scalable MVS algorithm.
Instead of using 3D CNN, some recent methods~\cite{yao2019recurrent,yan2020dense} apply recurrent convolution structure for cost volume regularization, which is effective and memory efficient to reconstruct scenes with wide ranges of depth.

To tackle the aforementioned problems, we therefore present a novel long short-term memory (LSTM) based recurrent multi-view stereo network with both intra-view and inter-view adaptive aggregation modules, namely AA-RMVSNet.
The intra-view scheme is designed for robust feature extraction, where context-aware features are adaptively aggregated for multiple scales and regions with varying richness of texture;
the inter-view scheme is used at multi-view cost volume aggregation step, whose aim is to overcome the difficulty of varying occlusion in complex scenarios by allocating higher weights on the well-matched view pairs.
As a result, the proposed network is able to obtain accurate and complete depth maps to further generate high quality dense point clouds, as illustrated in Fig.~\ref{first}.

The main contributions of this work are listed below:
\begin{itemize}
\setlength{\itemsep}{0pt}
\setlength{\parsep}{0pt}
\setlength{\parskip}{0pt}
\item We introduce an intra-view feature aggregation module to adaptively extract image features by using deformable convolution and multi-scale aggregation.
\item We propose an inter-view cost volume aggregation module to adaptively aggregate cost volumes of different views by yielding pixel-wise attention maps for each view.
\item Our method ranks $1^{st}$ among all submissions on Tanks and Temples online benchmark and obtains competitive results on DTU dataset.
\end{itemize}

\section{Related Work}
\subsection{Traditional MVS}

According to output scene representations, traditional MVS reconstruction methods can be categorized into three types: volumetric~\cite{kutulakos2000theory,seitz1999photorealistic}, point-based~\cite{lhuillier2005quasi,furukawa2009accurate} and depth-based~\cite{campbell2008using,galliani2015massively,schonberger2016mvs,xu2019acmm}. Volumetric methods first discretize the whole 3D space into regular cubes and then decide whether a voxel belongs to the surface or not with the photometric consistency metric. 
The space discretization is memory intensive, thus these methods are not scalable to large-scale scenarios. 
Point-based methods focus on the 3D points, usually start from a sparse set of matched key points and use the propagation strategy to gradually densify the reconstruction, which limits the capacity of parallel data processing. 
In contrast, depth-based methods have shown more flexibility in modeling the 3D geometry of scene. 
It reduces the MVS reconstruction into relatively small problems of per-view depth map estimation, and can be further fused to point cloud~\cite{merrell2007real} or the volumetric reconstructions~\cite{newcombe2011kinectfusion}. 
Many successful traditional MVS algorithms yielding depth maps have been proposed. Sch{\"o}nberger \etal present COLMAP~\cite{schonberger2016mvs}, which uses hand-crafted features and jointly estimates pixel-wise view selection, depth map and surface normal to utilize the photometric and geometric priors. Xu \etal propose ACMM~\cite{xu2019acmm} with multi-scale geometric consistency, adaptive checkerboard sampling and multi-hypothesis joint view selection. Although traditional MVS methods yield impressive results, they utilize hand-crafted features which are not suitable for non-Lambertian surfaces, low-textured and texture-less regions where photometric consistency is unreliable. 

\subsection{Learning-based MVS} 
\begin{figure*}[t]
    \centering
    \includegraphics[width=\linewidth]{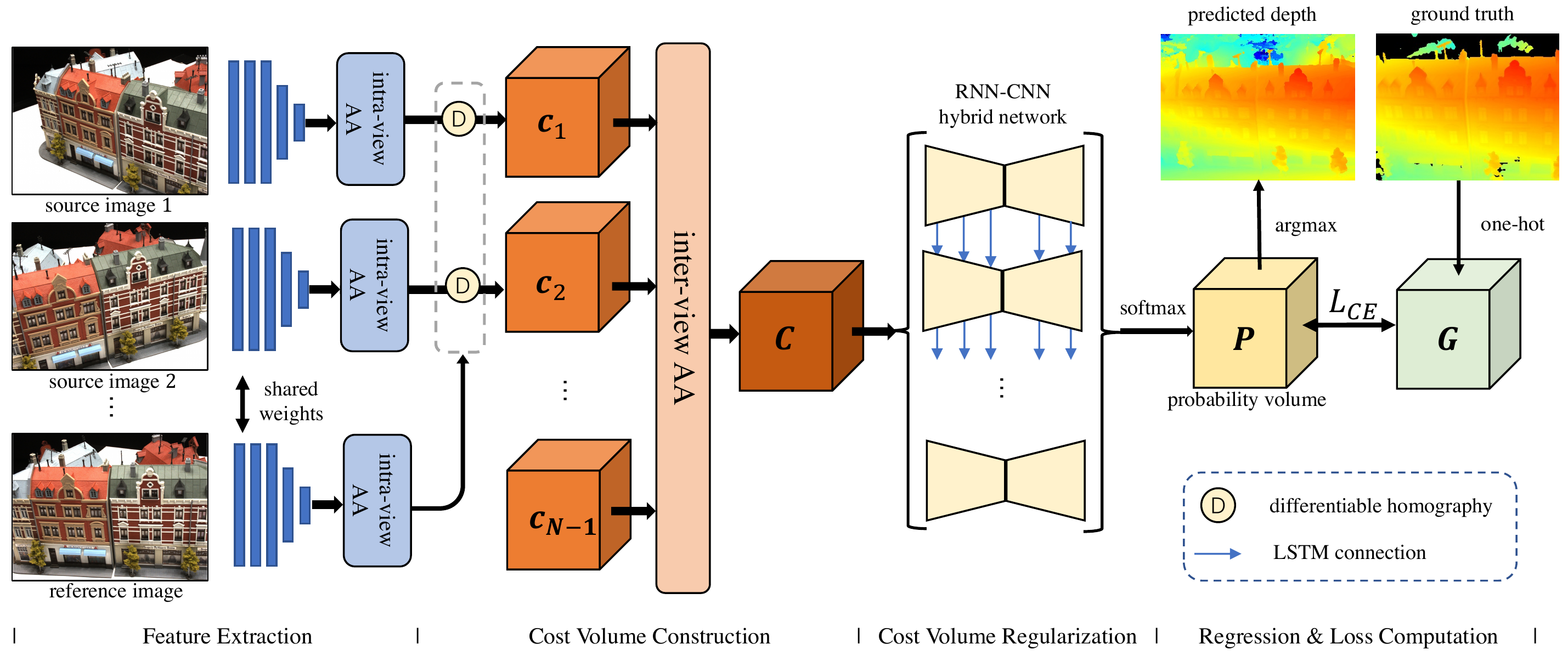}
    \caption{Overall architecture of AA-RMVSNet that consists of 4 stages. Intra-view AA module aims to aggregate context-aware features for multiple scales and regions with varying richness of texture. Inter-view AA module adaptively aggregates cost volumes of different views by yielding pixel-wise attention maps for each view. A RNN-CNN hybrid network is adopted to regularize cost volumes in a recurrent slice-by-slice pattern. At last, cross entropy for pixel-wise classification is adopted to calculate the loss for back propagation.}
    \label{fig:overall}
\end{figure*}
Rather than using traditional hand-crafted image features, recent studies on MVS apply deep learning for better reconstruction accuracy and completeness. Volumetric methods SurfaceNet~\cite{ji2017surfacenet} and LSM~\cite{kar2017learning} are first proposed. They construct a cost volume using multi-view images and use 3D CNNs to regularize and infer the voxel. However, SurfaceNet and LSM are restricted to only small-scale reconstructions due to the common drawback of the volumetric representation. Depth-based method MVSNet~\cite{yao2018mvsnet} improves the MVS reconstruction performance a lot compared with SurfaceNet and LSM. MVSNet takes one reference image and several source images as input and extracts deep image features, then encodes camera geometries in the network to build the 3D cost volumes via differentiable homography. To reduce the huge memory consumption of MVSNet, some variants of MVSNet have been proposed recently and can be divided into: multi-stage methods and recurrent methods. Multi-stage methods, such as CasMVSNet~\cite{gu2020cascade}, CVP-MVSNet~\cite{yang2020cost}, UCS-Net~\cite{cheng2020deep}, Vis-MVSNet~\cite{zhang2020visibility}, use the coarse-to-fine strategy that first predict a low resolution depth map with large depth interval and iteratively upsamples and refines the depth map with a narrow depth range. Though the coarse-to-fine architectures successfully reduce memory consumption, they are not suitable for high-resolution depth reconstructions as the depth prediction of coarse stage may be wrong with a large depth interval. To this end, recurrent methods, such as R-MVSNet~\cite{yao2019recurrent} and $D^2$HC-RMVSNet~\cite{yan2020dense}, are proposed. They sequentially regularize cost maps along the depth dimension with recurrent networks to avoid using memory-intensive 3D CNNs; thus they can infer depth maps within a very large depth range. R-MVSNet regularizes cost volumes in a sequential manner using convolutional gated recurrent unit (GRU). $D^2$HC-RMVSNet improves R-MVSNet with more powerful recurrent convolutional cells, ConvLSTMCells, and a dynamic consistency checking strategy. 

Though achieving promising results, most of the aforementioned learning-based MVS methods still have difficulties in handling challenging regions and severe occlusion problems in MVS. 

\section{Methodology}
The overall architecture of our AA-RMVSNet follows the typical pattern of a learning-based MVS pipeline, which is illustrated in Fig.~\ref{fig:overall}. Input images are separated into $1$ reference image and $N-1$ source images. Image features ($H\times W\times F$) of all $N$ images are extracted by an encoder with shared weights and a 3D cost volume ($H\times W\times D\times F$) is constructed via differentiable homography by warping features of source images to the reference camera frustum. Then the cost volume is regularized to obtain a probability volume $H\times W\times D$ which generates the prediction of depth map. Feature maps of all images are filtered and fused to obtain the dense point cloud of the scene.

Particularly for AA-RMVSNet, per-view matching cost volumes are computed by matching features of the $N-1$ warped source images and the reference image with $D$ depth hypotheses. 
The pixel-wise mapping relation between the reference image and the $i$-th source image with depth hypothesis $d$ is described by the differentiable homography as
\begin{equation}
    \mathbf{H}_i^{(d)} = d\mathbf{K}_i\mathbf{T}_i\mathbf{T}^{-1}_{ref}\mathbf{K}^{-1}_{ref},
\end{equation}
where $\mathbf{T}$ and $\mathbf{K}$ denote camera extrinsics and intrinsics respectively. Then per-view cost volumes are calculated by
\begin{equation}
    \mathbf{c}_i^{(d)} = (\mathbf{f}_{src_i}^{(d)} - \mathbf{f}_{ref})^2,
\end{equation}
where $\mathbf{f}_{src_i}^{(d)}$ represents extracted features of the $i$-th source image and $\mathbf{f}_{ref}$ represents features of the reference image.
All $N-1$ cost volumes are aggregated and cost volume regularization is then carried out by a hybrid neural network to obtain depth maps and corresponding probability distribution.

AA-RMVSNet further improves the pipeline by leveraging the idea of adaptive aggregation at two stages, namely intra-view adaptive aggregation (intra-view AA) at feature extraction (Sec.~\ref{sec:intra}) and inter-view adaptive aggregation (inter-view AA) at cost volume construction (Sec.~\ref{sec:inter}).
Besides, a RNN-CNN hybrid neural network (Sec.~\ref{sec:lstm}), which is memory efficient and robust for varying scenes, is adopted to commit cost volume regularization recurrently.

\subsection{Intra-view Adaptive Aggregation}
\label{sec:intra}
\begin{figure}[t]
    \centering
    \includegraphics[width=0.9\columnwidth]{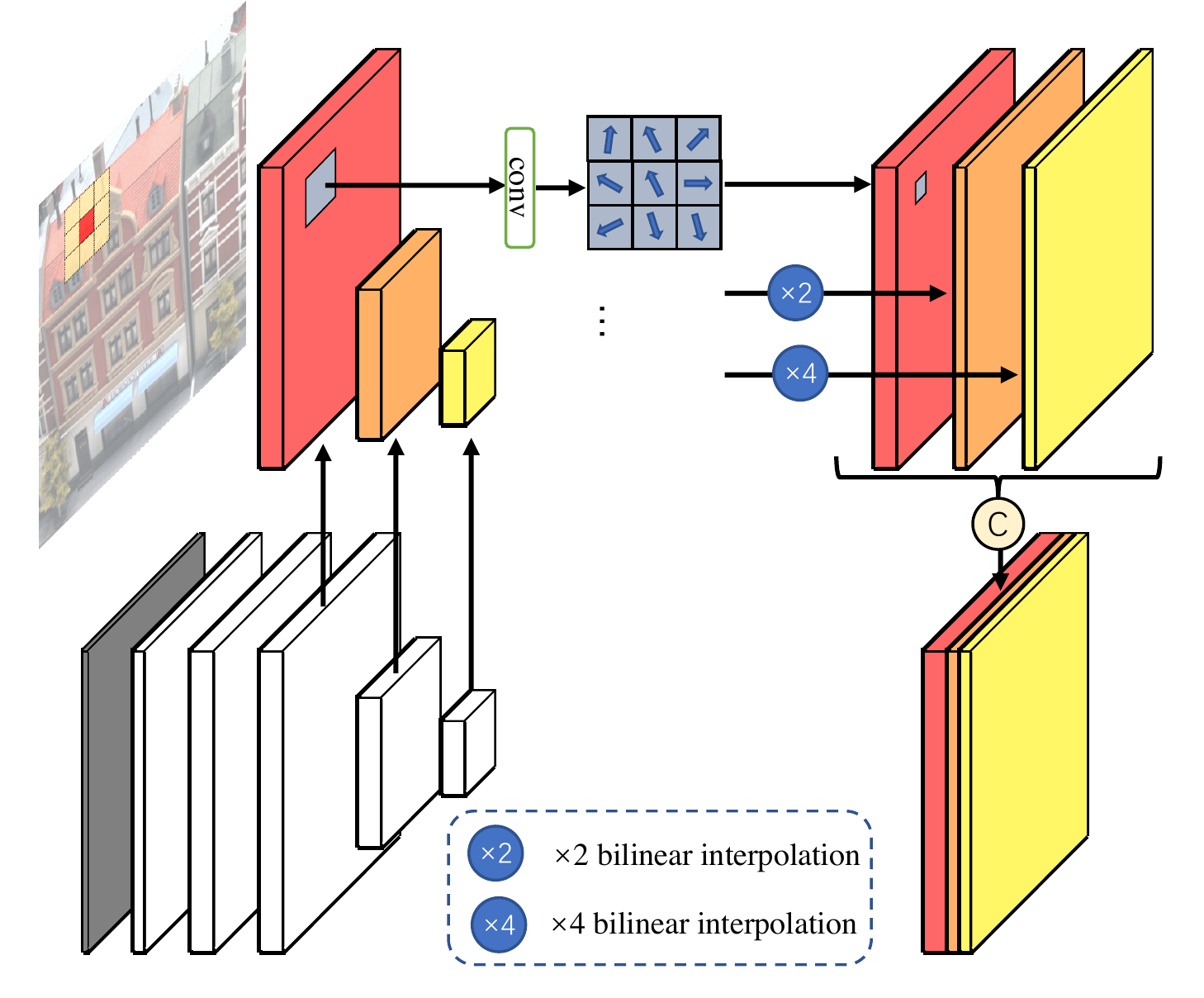}
    \caption{Intra-view AA module. All convolution kernels are $3\times 3$. Feature channels at the encoder (colored white) are 8, 16, 16, 16, 16. Multi-scale feature maps are sent into three deformable convolutions respectively, whose parameters are not shared. By bilinear interpolation and concatenation, a feature map of $H\times W\times (16+8+8)$ is built.}
    \label{fig:intra}
\end{figure}

As have been covered, 3D cost volumes are constructed by matching 2D feature maps so extracting recognizable and reliable features is of great significance in MVS. As for 3D reconstruction, it is universally acknowledged that reflective surfaces and low-textured or texture-less areas are main difficulties for a common CNN to handle which is operated on regular 2D grids with fixed receptive fields. For those challenging regions that are generally lacking in texture, we expect the receptive fields of convolutions to be larger while smaller receptive fields are favored for regions with rich texture. We introduce an intra-view AA module illustrated as Fig.~\ref{fig:intra} for adaptive aggregating features of different scales and regions with varying richness of texture. In the intra-view AA module, 3 feature maps of different spatial scales, whose sizes are $H\times W\times 16$, $\frac{H}{2}\times \frac{W}{2}\times 16$ and $\frac{H}{4}\times \frac{W}{4}\times 16$ respectively, are processed by 3 one-stride deformable convolutions~\cite{dai2017deformable,zhu2019deformable} with exclusive parameters. The definition of a deformable convolution is defined as
\begin{equation}
   \mathbf{f}'(\mathbf{p}) =\sum_{k}w_k \cdot\mathbf{f}(\mathbf{p}+\mathbf{p}_k+\Delta\mathbf{p}_k)\cdot \Delta m_k,
\end{equation}
where $\mathbf{f}(\mathbf{p})$ denotes the feature value pixel $\mathbf{p}$; $w_k$ and $\mathbf{p}_k$ represent the kernel parameter and fixed offset defined in a common convolution operation; $\Delta \mathbf{p}_k$ and $\Delta m_k$ are the offset and modulation weight yielded adaptively by learnable sub-networks of deformable convolution. By interpolating smaller feature maps to $H\times W$, we obtain 3 feature maps with 16, 8, 8 channels respectively and these features are concatenated to construct a feature map of $H\times W\times 32$.

\subsection{Inter-view Adaptive Aggregation}
\label{sec:inter}

After per-view cost volumes have been constructed, the next step is to aggregate all cost volumes into one for regularization.

A common practice is to average $N-1$ cost volumes, whose underlying principle is that all views should be of equal importance.
However, this is not reasonable enough since varying shooting angles may lead to problems such as occlusion and different lighting conditions of non-Lambertian surfaces that make depth estimation more difficult.

\begin{figure}[t]
    \centering
    \includegraphics[width=\columnwidth]{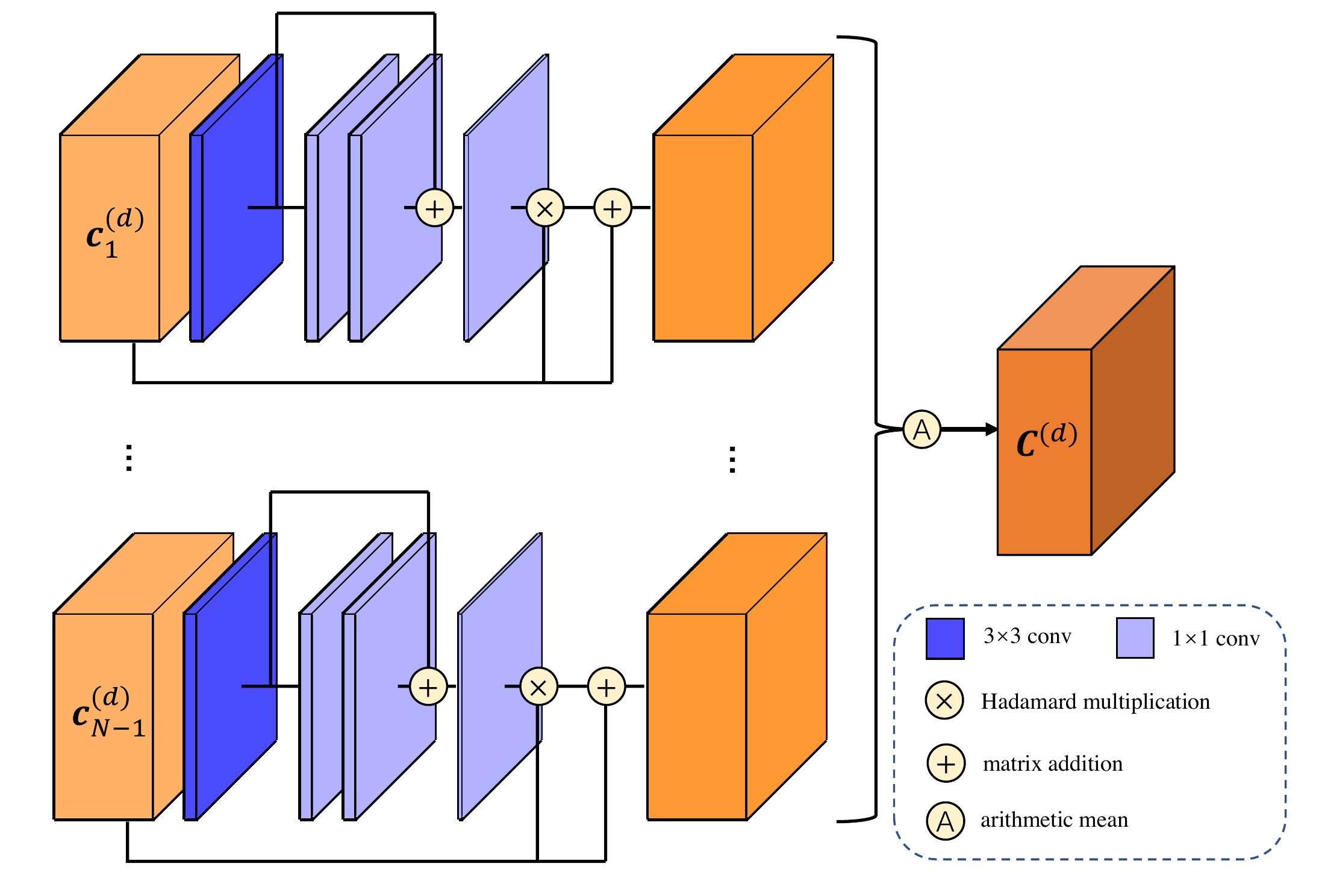}
    \caption{Inter-view AA module. For an input cost volume of $H\times W\times 32$, the following intermediate channel numbers are 4, 4, 4, 1. After reweighted by a $H\times W\times 1$ attention map, all cost volumes are summed and divided by $N-1$.}
    \label{fig:inter}
\end{figure}

Therefore, as illustrated in Fig.~\ref{fig:inter}, we design an inter-view AA module to handle unreliable matching costs, which is defined as 
\begin{equation}
    \mathbf{C}^{(d)} = \frac{1}{N-1}\sum_{i=1}^{N-1}[1+\boldsymbol{\omega}(\mathbf{c}_i^{(d)})] \odot \mathbf{c}_i^{(d)},
\end{equation}
where $\odot$ denotes Hadamard multiplication and  $\boldsymbol{\omega}(\cdot)$ is pixel-wise attention maps adaptively yielded according to per-view cost volumes. In this way, pixels that are likely to be confusing for matching will be suppressed while those with crucial context information will be assigned with larger weights. $1+\boldsymbol{\omega}(\cdot)$ better prevents over-smoothness than $\boldsymbol{\omega}(\cdot)$ alone.
\subsection{Recurrent Cost Regularization}
\label{sec:lstm}
Cost regularization is to leverage spatial context information and to turn matching costs into a probability distribution of $D$ depth hypotheses.
The regularization network adopts a RNN-CNN hybrid fashion where a cost volume ($H\times W\times D\times 32$) is sliced at the dimension of $D$. As is illustrated in Fig.~\ref{fig:overall}, feature passing in the regularization network has both horizontal direction and vertical direction. Horizontally, each slice of 3D cost volume is regularized by a CNN with encoder-decoder architecture; on the vertical direction, there are 5 parallel RNNs to deliver intermediate outputs of former ConvLSTMCells to later ones.

Considering a cost volume slice of depth hypothesis $d$ to be processed by the $j$-th convolution layer, denoted as $\mathbf{v}_{j-1}^{(d)}$, the output of this layer with depth hypothesis $d-1$ as $\mathbf{v}_{j}^{(d-1)}$ and memory maintained (or hidden state) as $\mathbf{m}_{j}^{(d-1)}$, operations within a ConvLSTMCell are as follows. 

Firstly, $\mathbf{v}_{j-1}^{(d)}$ and $\mathbf{v}_{j}^{(d-1)}$ are concatenated and after being processed by a convolution layer, the tensor is split into 4 tensors from the feature dimension, namely $\mathbf{w}$, $\mathbf{x}$, $\mathbf{y}$ and $\mathbf{z}$. The 4 signals within a LSTM cell are defined as
\begin{equation}
    \left\{
\begin{aligned}
\mathbf{i} & =  \sigma(\mathbf{w}) \\
\mathbf{f} & =  \sigma(\mathbf{x}) \\
\mathbf{o} & =  \sigma(\mathbf{y})\\
\mathbf{g} & =  \tanh(\mathbf{z}) 
\end{aligned}
\right.
\end{equation}
where all signals are two-dimensional in space and Sigmoid function $\sigma(\cdot)$ and hyperbolic tangent function $\tanh(\cdot)$ are all element-wise operations. Then the memory of LSTM is updated by
\begin{equation}
    \mathbf{m}_{j}^{(d)} = \mathbf{m}_{j}^{(d-1)} \odot \mathbf{f}+\mathbf{i}\odot \mathbf{g},
\end{equation}
while the output of the cell is 
\begin{equation}
    \mathbf{v}_{j}^{(d)} = \mathbf{o}\odot \tanh(\mathbf{m}_{j}^{(d)}).
\end{equation}

\subsection{Loss Function}
Since cost volume regularization turns matching costs into a pixel-wise probability distribution of depth hypothesis, the task of depth estimation is now similar to a pixel-wise classification problem. Therefore, by encoding the ground truth with one-hot pattern, we adopt cross entropy to calculate the training loss, defined as
\begin{equation}
    L = \sum_{\mathbf{p}\in \{\mathbf{p}_{v}\}}\sum_{d=d_0}^{d_{D-1}} -G^{(d)}(\mathbf{p})\ \log [P^{(d)}(\mathbf{p})],
\end{equation}
where $G^{(d)}(\mathbf{p})$ and $P^{(d)}(\mathbf{p})$ denote ground truth probability and predicted probability of depth hypothesis $d$ at pixel $\mathbf{p}$. $\{\mathbf{p}_{v}\}$ is the set of valid pixels with reliable depth.
\section{Experiments}

\begin{figure}[htbp]
    \centering
    \includegraphics[width=\columnwidth]{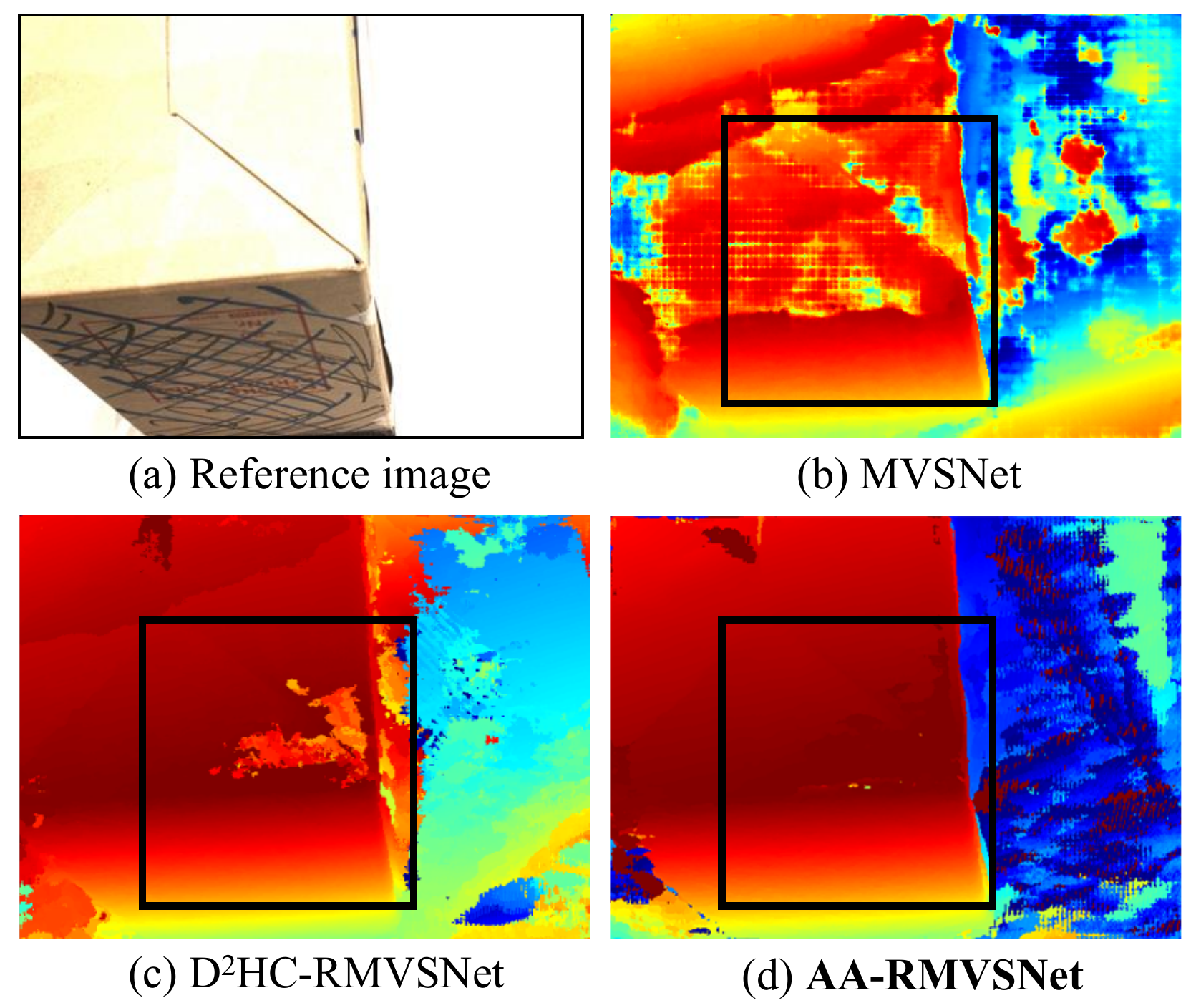}
    \caption{Comparison of depth map estimation of Scan 13 in DTU evaluation set~\cite{Aans2016Large}. Our AA-RMVSNet's prediction is more accurate, continuous and complete in contrast to~\cite{yao2018mvsnet,yan2020dense}.}
    \label{dtu_depth}
\end{figure}

\subsection{Datasets}
\paragraph{DTU dataset}
DTU dataset~\cite{Aans2016Large} is an indoor MVS dataset collected under well-controlled laboratory conditions with fixed camera trajectory. It contains 128 scans with 49 views under 7 different lighting conditions and is split into 79 training scans, 18 validation scans and 22 evaluation scans. By setting each image as reference, there are 27097 training samples in total. Following common configurations, we apply DTU dataset for network training and evaluation.

\paragraph{BlendedMVS dataset}
BlendedMVS dataset~\cite{yao2020blendedmvs} is a recently published large-scale synthetic dataset for multi-view stereo training containing a variety of indoor and outdoor scenes, such as cities, architectures, sculptures and shoes. The dataset consists of over 17k high-resolution images and is split into 106 training scenes and 7 validation scenes. However, this dataset does not officially provide evaluation tools, so we utilize BlendedMVS dataset for network fine-tuning and qualitative evaluation. 

\paragraph{Tanks and Temples benchmark}
Tanks and Temples~\cite{knapitsch2017tanks} is a large-scale outdoor benchmark captured in more complex real scenarios. It contains an intermediate set and an advanced set. Specifically, the intermediate set has eight scenes: Family, Francis, Horse, Lighthouse, M60, Panther, Playground and Train. Different scenes have different scales, surface reflection and exposure conditions. Evaluation of Tanks and Temples benchmark is done online by uploading reconstructed point clouds to its official website~\cite{tnt}.
Until now, there have been hundreds of submissions on Tanks and Temples leaderboard including almost all recent state-of-the-art methods.

\begin{figure*}[t]
    \centering
    \includegraphics[width=2\columnwidth]{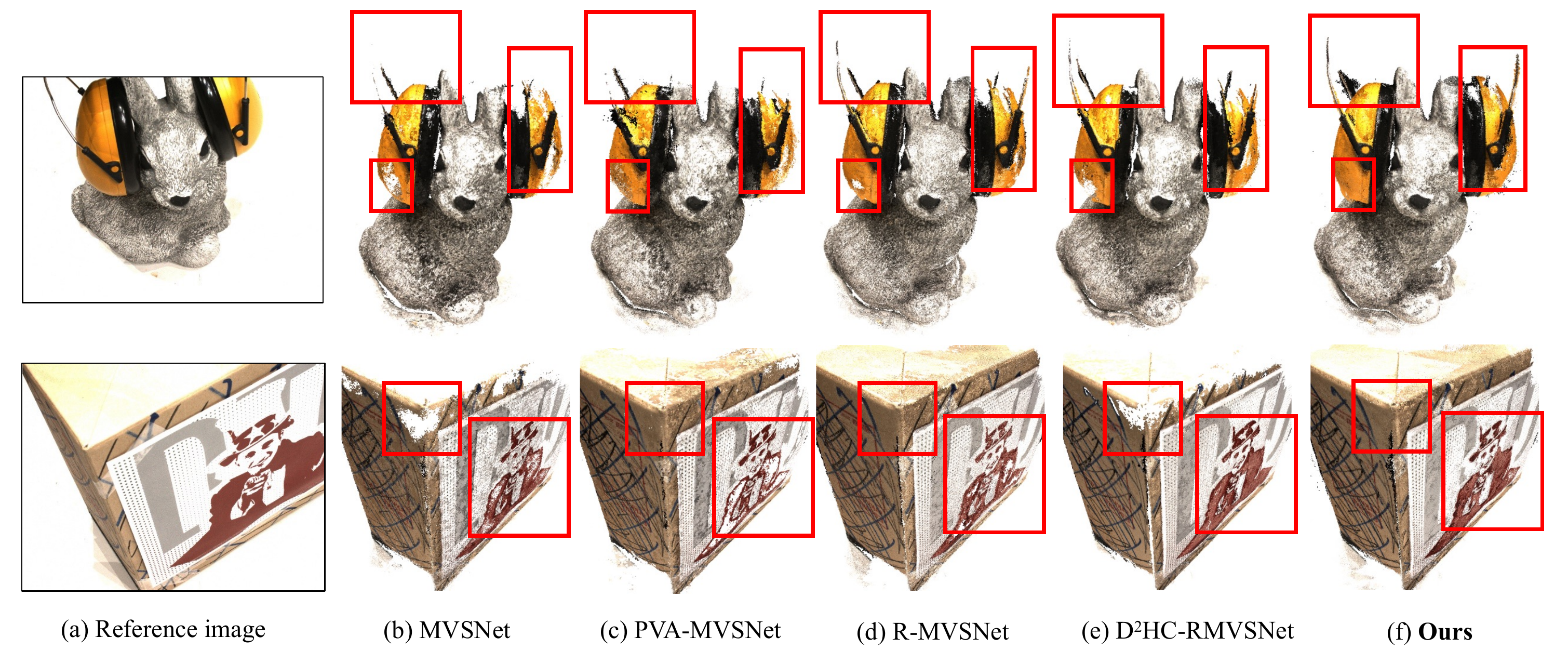}
    \caption{Qualitative comparisons with~\cite{yao2018mvsnet,yi2020PVAMVSNET,yao2019recurrent,yan2020dense} of Scan 33 and Scan 13 in DTU dataset~\cite{Aans2016Large}. Our method provides more complete 3D dense point clouds with details preserved.}
    \label{dtu_point}
\end{figure*}

\begin{table}[t]
    \footnotesize
    \centering
    \begin{tabular}{lccc}
    \hline
    Method & Acc.(mm) & Comp.(mm) & \textsl{Overall}(mm) \\
    \hline
    Furu \cite{furukawa2009accurate} & 0.613 & 0.941 & 0.777 \\ 
     Gipuma~\cite{galliani2015massively} & \textbf{0.283} & 0.873 & 0.578 \\
    COLMAP~\cite{schonberger2016mvs} & 0.400 & 0.664 & 0.532 \\ 
    MVSNet~\cite{yao2018mvsnet} & 0.396 & 0.527 & 0.462 \\ 
    R-MVSNet~\cite{yao2019recurrent} & 0.385 & 0.459 & 0.422 \\ 
    P-MVSNet~\cite{luo2019p} & 0.406 & 0.434 & 0.420 \\
    PointMVSNet~\cite{chen2019point} & 0.361 & 0.421 & 0.391 \\ 
    $D^2$HC-RMVSNet~\cite{yan2020dense} & 0.395 & 0.378 & 0.386 \\
    PointMVSNet~\cite{chen2019point} & 0.342 & 0.411 & 0.376 \\ 
    Vis-MVSNet~\cite{zhang2020visibility} & 0.369  & 0.361 & 0.365 \\
    CasMVSNet~\cite{gu2020cascade} & 0.325 &  0.385 & 0.355 \\ 
    CVP-MVSNet~\cite{yang2020cost} & 0.296 &  0.406 & \textbf{0.351} \\ 
     \hline
    \rule{0pt}{9pt}
    \textbf{AA-RMVSNet} & 0.376 & \textbf{0.339} & 0.357 \\[2pt]
   \hline
    \end{tabular}
    \caption{Quantitative results on DTU evaluation set~\cite{Aans2016Large} (lower is better). Our method AA-RMVSNet exhibits a competitive \textsl{overall} score compared with other state-of-the-art methods. Specially, our method outperforms all methods mentioned in terms of \textsl{completeness}.}
    \label{eval_dtu}
\end{table}

\subsection{Implementation Details}
\label{Implementation}
\paragraph{Training}
We train our AA-RMVSNet on DTU training set~\cite{Aans2016Large} consisting of 79 different scenes.
Since DTU dataset only provides laser ground truth point clouds, in order to obtain ground truth depth maps for network training, we follow the previous MVS methods~\cite{yao2018mvsnet,yao2019recurrent,zhang2020visibility,gu2020cascade} to generate coarse ground truth depth maps by screened Poisson surface reconstruction algorithm~\cite{2013screened} and depth rendering.
After that, we improve the reliability of original depth maps by cross-filtering with their neighboring views, which is similar to~\cite{2020attention}.
We resize the original images to the size of $W\times H=160\times 128$ which is equal to the resolution of the refined ground truth depth maps.
The number of input images is set to $N=7$ while the number of depth hypotheses is set to $D=192$, which is uniformly sampled from $425mm$ to $935mm$.
We implement our AA-RMVSNet by PyTorch~\cite{paszke2017automatic} and train
the proposed network end-to-end using Adam~\cite{kingma2014adam} with an initial learning rate of 0.001, which decays by 0.9 each epoch.
The total training phase costs 20.16GB memory and takes about 3 days.
Batch size is set to 4 on 4 NVIDIA TITAN RTX GPUs.

\paragraph{Testing}

Since the training phase needs extra memory to save intermediate gradients for back propagation, the testing phase of AA-RMVSNet is relatively memory efficient so that it could deal with higher resolution images and finer depth plane sweep.
We set $N=7$ and $D=512$ in the testing phase to obtain depth maps with finer details.
In order to fit the network, the height and width of input images must be a multiple of 8.
We use input images of $800\times 600$ resolution for DTU evaluation.
Before testing on BlendedMVS, we fine-tune our network on the training set of BlendedMVS to boost the performance of various scenarios.
We test our network on the validation set of BlendedMVS using original images of $768\times 576$ with inverse depth setting.
For benchmarking on Tanks and Temples, we apply COLMAP-SfM~\cite{schonberger2016sfm} to estimate depth ranges and camera parameters.
Different from the image cropping methods in~\cite{yao2018mvsnet,yao2019recurrent,yi2020PVAMVSNET,yan2020dense,gu2020cascade}, we resize and pad images to the size of $1024\times 544$ or $960\times 544$ to fit our network, so context information near image boundary is preserved in this way.

\paragraph{Filtering and Fusion}

Similar to the previous MVS methods~\cite{yao2018mvsnet,yao2019recurrent,yi2020PVAMVSNET,gu2020cascade,2020attention}, we introduce photometric and geometric constraints for depth map filtering.
The photometric constraint measures the multi-view matching quality, where depth with low confidence value is considered as an outlier.
In our experiments, we discard pixels whose probability of estimated depth is lower than 0.3. 
The geometric constraint measures multi-view depth consistency, where depth inconsistent with its neighboring views should also be discarded.
We follow the dynamic geometric consistency checking method presented in~\cite{yan2020dense} to cross-filter original depth maps.
After that, we utilize a visibility-based depth fusion method proposed by~\cite{merrell2007real} with a mean average fusion approach~\cite{yao2018mvsnet} to produce final 3D point clouds.

\begin{table*}[htbp]
    \centering
    \begin{tabular}{lcccccccccc}
    \hline
     \rule{0pt}{13pt} 
    Method & \textbf{Rank} & \textbf{Mean} & Family & Francis & Horse & L.H. & M60 & Panther & P.G. & Train \\[5pt]
    \hline
    CIDER~\cite{xu2020cider}&95.00 &	46.76&	56.79&	32.39&	29.89&	54.67&	53.46&	53.51&	50.48&	42.85\\
   Point-MVSNet~\cite{chen2019point}&93.88&	48.27	&	61.79&	41.15&	34.20	&50.79&	51.97&	50.85&	52.38	&43.06\\
  Dense R-MVSNet~\cite{yao2019recurrent}&83.50&	50.55&	73.01&	54.46&	43.42&	43.88	&46.80&	46.69&	50.87&	45.25\\
      PVA-MVSNet~\cite{yi2020PVAMVSNET}&56.62&  54.46	&	69.36 & 46.80 &	46.01 &	55.74 &	57.23 &	54.75 &	56.70 &	49.06 \\
    CVP-MVSNet~\cite{yang2020cost} &55.12&	54.03&	76.50&	47.74&	36.34&	55.12&	57.28&	54.28&	57.43&	47.54\\
  P-MVSNet~\cite{luo2019p}&	43.12 & 55.62&	70.04&	44.64&	40.22&	{65.20}&	55.08&	55.17&	60.37&	54.29 \\
    CasMVSNet~\cite{gu2020cascade}&40.38	&56.84	&	76.37&	58.45&	46.26&	55.81&	56.11&	54.06&	58.18&	49.51\\
    ACMM~\cite{xu2019acmm}&34.25	&57.27	&	69.24	&51.45	&46.97	&63.20&	55.07&	57.64&	60.08&	54.48 \\
    DeepC-MVS~\cite{kuhn2020deepc} &24.62&	59.79&	71.91&	54.08&	42.29&	\textbf{66.54}&	55.77&	\textbf{67.47}&	60.47&	\textbf{59.83}\\
    Altizure-HKUST-2019~\cite{altizure}&24.00&	59.03	&	{77.19} & {61.52} & 42.09 & 63.50& 59.36&	58.20&	57.05&	53.30\\
    AttMVS~\cite{2020attention} &19.00&	{60.05} &	73.90 &	\textbf{62.58} &	44.08 &	64.88 &	56.08&	59.39 &	\textbf{63.42} &	56.06\\
    $D^2$HC-RMVSNet~\cite{yan2020dense}&18.38	&	59.20&	74.69&	56.04&	49.42&	60.08&	59.81&{59.61}&	60.04&	53.92 \\
    Vis-MVSNet~\cite{zhang2020visibility}&15.38	&{60.03} &	77.40&	60.23&	47.07&	63.44&	{62.21}&	57.28&	{60.54}&	52.07\\
    \hline
    \rule{0pt}{12pt} 
    \textbf{AA-RMVSNet} &\textbf{6.38}&	\textbf{61.51}&\textbf{77.77}&	59.53&	\textbf{51.53}&	64.02&	\textbf{64.05}&	59.47&	60.85&	54.90\\[4pt]
    \hline
    \end{tabular}
    \caption{Benchmarking results on the Tanks and Temples~\cite{knapitsch2017tanks}. The evaluation metric is mean \textsl{F-score} (higher is better). AA-RMVSNet outperforms all existing MVS methods with a significant margin and ranks $1^{st}$ on Tanks and Temples leaderboard (Mar. 15, 2021). The \textbf{Rank} is a metric representing the average rank of all 8 scenes and is the basis for final ranking.}
    \label{tnt_eval}
\end{table*}

\begin{figure*}[t]
    \centering
    \includegraphics[width=2\columnwidth]{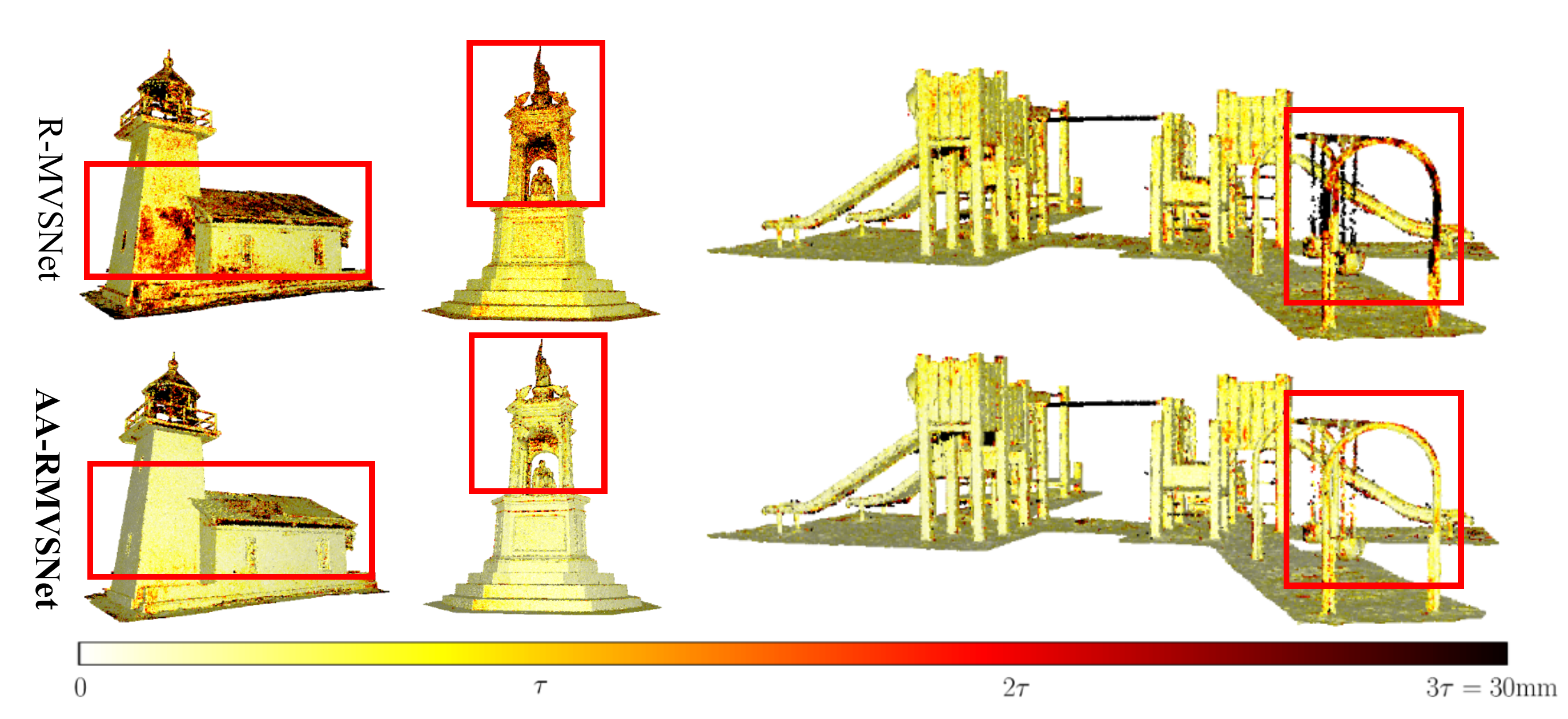}
    \caption{Error visualization of Lighthouse, Francis and Playground in Tanks and Temples benchmark~\cite{knapitsch2017tanks} calculated according to corresponding ground truth point clouds, in contrast to R-MVSNet~\cite{yao2019recurrent}.}
    \label{tnt}
\end{figure*}
\subsection{Experimental Results}

\paragraph{Results on DTU dataset}

We firstly evaluate AA-RMVSNet on DTU evaluation set~\cite{Aans2016Large}.
The depth comparison of Scan 13 with ~\cite{yao2018mvsnet,yan2020dense} is shown in Fig.~\ref{dtu_depth}.
Benefited from the intra-view AA module which integrates multi-scale and context-aware features, our method is able to estimate more complete and continuous depths for the low-textured surface of the paper box.
Some qualitative results compared with other methods are shown in Fig.~\ref{dtu_point}. Due to the improvement of depth map estimation, our method obtains more complete 3D dense point clouds with details reserved.
The quantitative results of the whole DTU evaluation set are shown in Tab.~\ref{eval_dtu}, where 
\textsl{accuracy} and \textsl{completeness} are two absolute distances calculated by the official MATLAB evaluation code~\cite{Aans2016Large}. 
\textsl{Overall} is the mean average of the two metrics.
Compared with the advanced methods, our method achieves best \textsl{completeness} and competitive \textsl{overall} performance.
Through the comparison with two previous recurrent MVS networks R-MVSNet and $D^2$HC-RMVSNet, our method significantly improves both \textsl{accuracy} and \textsl{completeness} on DTU dataset.

\paragraph{Benchmarking on Tanks and Temples}

In order to evaluate the performance of our method under complex outdoor scenes, we test our method on Tanks and Temples benchmark as demonstrated in Tab.~\ref{tnt_eval}.
Our proposed AA-RMVSNet outperforms all existing MVS methods with a significant margin and ranks $1^{st}$ on Tanks and Temples leardboard (Mar. 15, 2021) with 61.51 mean \textsl{F-score}.
Compared with the state-of-the-art methods on DTU dataset, such as CasMVSNet and CVP-MVSNet, our method exhibits stronger robustness and generalizability for varying scenarios.
Fig.~\ref{tnt} visualizes the error maps calculated according to the corresponding ground truth point clouds.
In contrast to the original recurrent MVS network R-MVSNet, our method significantly improves overall reconstruction quality, especially at challenging regions such as low-textured planes, occluded areas and thin objects, which is benefited from our robust feature extraction and view aggregation methods.

\paragraph{Results on BlendedMVS dataset}
To further demonstrate the generalizability and scalablility of our method, we also test it on BlendedMVS validation set~\cite{yao2020blendedmvs}. 
Our method successfully reconstructs whole wide-range aerial scenes as well as the small objects.
Please check the appendices for results.

\begin{figure}[t]
    \centering
    \includegraphics[width=0.9\columnwidth]{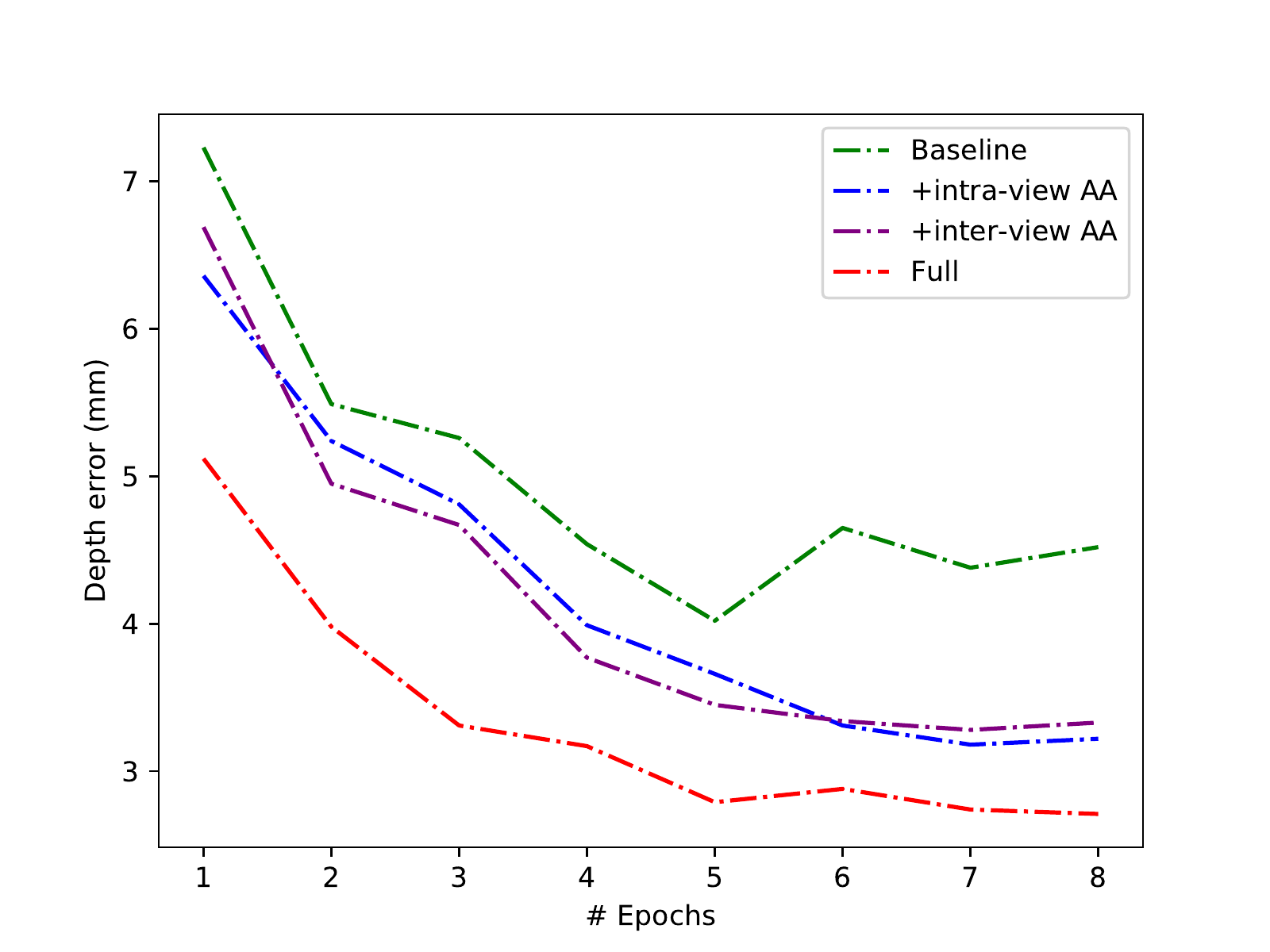}
    \caption{Validation results of the mean average depth error with different network architectures during training.}
    \label{training}
\end{figure}

\subsection{Ablation Study}

In this section, we provide ablation experiments to quantitatively analyze the effectiveness and memory cost of each adaptive aggregation method.
The following ablation studies are performed on DTU dataset using the same parameters as Sec.~\ref{Implementation}.
We compare four different network architectures with or without the proposed adaptive aggregation modules.
\textsl{Baseline} applies general 2D CNN for feature extraction and the same hybrid LSTM structure for cost volume regularization without any additional modules.

Validation results of the mean average depth error with different components during training are shown in Fig.~\ref{training}.
It is clear that each individual module can significantly lower the depth error, and the two modules are complementary in full AA-RMVSNet to achieve the best performance.

We also test the point cloud results generated by different network models as shown in Tab.~\ref{ablation}.
Both intra-view AA and inter-view AA can improve the accuracy and completeness of 3D reconstruction results.
Specifically, intra-view AA takes about 1.74GB additional memory and improves \textsl{completeness} by 0.28, while inter-view AA only costs extra 0.11 GB and gains 0.31 more in \textsl{accuracy}.
The \textsl{overall} error drops from 0.391 to 0.357 with both two modules.
Full AA-RMVSNet only takes 4.25GB to obtain dense and accurate depth maps with $800\times 600$ resolution, indicating that our method is fairly memory efficient.

Regarding ablation study for different experiment settings, please refer to the appendices for detailed results.

\begin{table}[t]
    \centering
    \footnotesize
    \begin{tabular}{lcccc}
    \hline
    Model  & Acc. & Comp. & \textsl{O.A.}(mm) & Mem.(GB) \\ 
    \hline
    Baseline & 0.408 & 0.374 & 0.391  & \textbf{2.41} \\ 
    +intra-view AA  & 0.396 & 0.346 & 0.371 & 4.15 \\
    +inter-view AA  & 0.377 & 0.363 & 0.370 & 2.52\\ 
     \hline
    \textbf{Full} & \textbf{0.376} & \textbf{0.339} & \textbf{0.357} & 4.25\\
   \hline
    \hline
     MVSNet~\cite{yao2018mvsnet} & 0.396 & 0.527 & 0.462  & 15.4\\
      R-MVSNet~\cite{yao2019recurrent} & 0.385 & 0.459 & 0.422  & 6.7\\
    \hline
    \end{tabular}
    \caption{Quantitative and memory performance with different components on DTU evaluation dataset~\cite{Aans2016Large}. }
    \label{ablation}
\end{table}


\section{Conclusion}

We have presented a novel recurrent multi-view stereo network with adaptive aggregation modules, denoted as AA-RMVSNet.
The intra-view feature aggregation module efficiently improves the performance on thin objects and large low-textured surfaces, by integrating multi-scale and context-aware features adaptively.
The inter-view cost volume aggregation module successfully handles the problem of varying occlusion in complex scenes by adaptive pixel-wise view aggregation.
The two modules are lightweight, effective and complementary.
As a result, our method achieves competitive results on DTU dataset and outperforms other submissions with a significant margin on Tanks and Temples benchmark, showing great generalizability and scalability.

\section*{Acknowledgements}
This research is supported by National Key Technology Research and Development Program of China, grant number 2017YFB1002601; National Natural Science Foundation of China (NSFC), grant number 61632003; PKU-Baidu Fund, grant number 2019BD007.


{\small
\bibliographystyle{ieee_fullname}
\bibliography{egbib}

\begin{thebibliography}{10}\itemsep=-1pt

\bibitem{altizure}
Altizure.
\newblock \url{https://github.com/altizure}.

\bibitem{tnt}
Tanks and temples benchmark.
\newblock \url{https://www.tanksandtemples.org}.

\bibitem{Aans2016Large}
Henrik Aan{\ae}s, Rasmus~Ramsb{\o}l Jensen, George Vogiatzis, Engin Tola, and
  Anders~Bjorholm Dahl.
\newblock Large-scale data for multiple-view stereopsis.
\newblock {\em International Journal of Computer Vision}, 120(2):153--168,
  2016.

\bibitem{barnes2009patchmatch}
Connelly Barnes, Eli Shechtman, Adam Finkelstein, and Dan~B Goldman.
\newblock Patchmatch: A randomized correspondence algorithm for structural
  image editing.
\newblock {\em ACM Trans. Graph.}, 28(3):24, 2009.

\bibitem{campbell2008using}
Neill~DF Campbell, George Vogiatzis, Carlos Hern{\'a}ndez, and Roberto Cipolla.
\newblock Using multiple hypotheses to improve depth-maps for multi-view
  stereo.
\newblock In {\em European Conference on Computer Vision}, pages 766--779.
  Springer, 2008.

\bibitem{chen2019point}
Rui Chen, Songfang Han, Jing Xu, and Hao Su.
\newblock Point-based multi-view stereo network.
\newblock In {\em Proceedings of the IEEE/CVF International Conference on
  Computer Vision}, pages 1538--1547, 2019.

\bibitem{cheng2020deep}
Shuo Cheng, Zexiang Xu, Shilin Zhu, Zhuwen Li, Li~Erran Li, Ravi Ramamoorthi,
  and Hao Su.
\newblock Deep stereo using adaptive thin volume representation with
  uncertainty awareness.
\newblock In {\em Proceedings of the IEEE/CVF Conference on Computer Vision and
  Pattern Recognition}, pages 2524--2534, 2020.

\bibitem{dai2017deformable}
Jifeng Dai, Haozhi Qi, Yuwen Xiong, Yi Li, Guodong Zhang, Han Hu, and Yichen
  Wei.
\newblock Deformable convolutional networks.
\newblock In {\em Proceedings of the IEEE international conference on computer
  vision}, pages 764--773, 2017.

\bibitem{furukawa2009accurate}
Yasutaka Furukawa and Jean Ponce.
\newblock Accurate, dense, and robust multiview stereopsis.
\newblock {\em IEEE transactions on pattern analysis and machine intelligence},
  32(8):1362--1376, 2009.

\bibitem{galliani2015massively}
Silvano Galliani, Katrin Lasinger, and Konrad Schindler.
\newblock Massively parallel multiview stereopsis by surface normal diffusion.
\newblock In {\em Proceedings of the IEEE International Conference on Computer
  Vision}, pages 873--881, 2015.

\bibitem{gu2020cascade}
Xiaodong Gu, Zhiwen Fan, Siyu Zhu, Zuozhuo Dai, Feitong Tan, and Ping Tan.
\newblock Cascade cost volume for high-resolution multi-view stereo and stereo
  matching.
\newblock In {\em Proceedings of the IEEE/CVF Conference on Computer Vision and
  Pattern Recognition}, pages 2495--2504, 2020.

\bibitem{ji2017surfacenet}
Mengqi Ji, Juergen Gall, Haitian Zheng, Yebin Liu, and Lu Fang.
\newblock Surfacenet: An end-to-end 3d neural network for multiview stereopsis.
\newblock In {\em Proceedings of the IEEE International Conference on Computer
  Vision}, pages 2307--2315, 2017.

\bibitem{kar2017learning}
Abhishek Kar, Christian H{\"a}ne, and Jitendra Malik.
\newblock Learning a multi-view stereo machine.
\newblock In {\em NIPS}, 2017.

\bibitem{2013screened}
Michael Kazhdan and Hugues Hoppe.
\newblock Screened poisson surface reconstruction.
\newblock {\em ACM Transactions on Graphics (ToG)}, 32(3):1--13, 2013.

\bibitem{kingma2014adam}
Diederik~P Kingma and Jimmy Ba.
\newblock Adam: A method for stochastic optimization.
\newblock {\em International Conference on Learning Representations (ICLR)},
  2014.

\bibitem{knapitsch2017tanks}
Arno Knapitsch, Jaesik Park, Qian-Yi Zhou, and Vladlen Koltun.
\newblock Tanks and temples: Benchmarking large-scale scene reconstruction.
\newblock {\em ACM Transactions on Graphics (ToG)}, 36(4):1--13, 2017.

\bibitem{kuhn2020deepc}
Andreas Kuhn, Christian Sormann, Mattia Rossi, Oliver Erdler, and Friedrich
  Fraundorfer.
\newblock Deepc-mvs: Deep confidence prediction for multi-view stereo
  reconstruction.
\newblock In {\em 2020 International Conference on 3D Vision (3DV)}, pages
  404--413. IEEE, 2020.

\bibitem{kutulakos2000theory}
Kiriakos~N Kutulakos and Steven~M Seitz.
\newblock A theory of shape by space carving.
\newblock {\em International journal of computer vision}, 38(3):199--218, 2000.

\bibitem{lhuillier2005quasi}
Maxime Lhuillier and Long Quan.
\newblock A quasi-dense approach to surface reconstruction from uncalibrated
  images.
\newblock {\em IEEE transactions on pattern analysis and machine intelligence},
  27(3):418--433, 2005.

\bibitem{luo2019p}
Keyang Luo, Tao Guan, Lili Ju, Haipeng Huang, and Yawei Luo.
\newblock P-mvsnet: Learning patch-wise matching confidence aggregation for
  multi-view stereo.
\newblock In {\em Proceedings of the IEEE/CVF International Conference on
  Computer Vision}, pages 10452--10461, 2019.

\bibitem{2020attention}
Keyang Luo, Tao Guan, Lili Ju, Yuesong Wang, Zhuo Chen, and Yawei Luo.
\newblock Attention-aware multi-view stereo.
\newblock In {\em Proceedings of the IEEE/CVF Conference on Computer Vision and
  Pattern Recognition}, pages 1590--1599, 2020.

\bibitem{merrell2007real}
Paul Merrell, Amir Akbarzadeh, Liang Wang, Philippos Mordohai, Jan-Michael
  Frahm, Ruigang Yang, David Nist{\'e}r, and Marc Pollefeys.
\newblock Real-time visibility-based fusion of depth maps.
\newblock In {\em 2007 IEEE 11th International Conference on Computer Vision},
  pages 1--8. IEEE, 2007.

\bibitem{newcombe2011kinectfusion}
Richard~A Newcombe, Shahram Izadi, Otmar Hilliges, David Molyneaux, David Kim,
  Andrew~J Davison, Pushmeet Kohi, Jamie Shotton, Steve Hodges, and Andrew
  Fitzgibbon.
\newblock Kinectfusion: Real-time dense surface mapping and tracking.
\newblock In {\em 2011 10th IEEE international symposium on mixed and augmented
  reality}, pages 127--136. IEEE, 2011.

\bibitem{paszke2017automatic}
Adam Paszke, Sam Gross, Soumith Chintala, Gregory Chanan, Edward Yang, Zachary
  DeVito, Zeming Lin, Alban Desmaison, Luca Antiga, and Adam Lerer.
\newblock Automatic differentiation in {PyTorch}.
\newblock {\em NeurIPS Autodiff Workshop}, 2017.

\bibitem{schonberger2016sfm}
Johannes~L Sch{\"o}nberger and Jan-Michael Frahm.
\newblock Structure-from-motion revisited.
\newblock In {\em Proceedings of the IEEE conference on computer vision and
  pattern recognition}, pages 4104--4113, 2016.

\bibitem{schonberger2016mvs}
Johannes~L Sch{\"o}nberger, Enliang Zheng, Jan-Michael Frahm, and Marc
  Pollefeys.
\newblock Pixelwise view selection for unstructured multi-view stereo.
\newblock In {\em European Conference on Computer Vision}, pages 501--518.
  Springer, 2016.

\bibitem{schops2017multi}
Thomas Schops, Johannes~L Sch{\"o}nberger, Silvano Galliani, Torsten Sattler,
  Konrad Schindler, Marc Pollefeys, and Andreas Geiger.
\newblock A multi-view stereo benchmark with high-resolution images and
  multi-camera videos.
\newblock In {\em Proceedings of the IEEE Conference on Computer Vision and
  Pattern Recognition}, pages 3260--3269, 2017.

\bibitem{seitz1999photorealistic}
Steven~M Seitz and Charles~R Dyer.
\newblock Photorealistic scene reconstruction by voxel coloring.
\newblock {\em International Journal of Computer Vision}, 35(2):151--173, 1999.

\bibitem{xu2019acmm}
Qingshan Xu and Wenbing Tao.
\newblock Multi-scale geometric consistency guided multi-view stereo.
\newblock In {\em Proceedings of the IEEE/CVF Conference on Computer Vision and
  Pattern Recognition}, pages 5483--5492, 2019.

\bibitem{xu2020cider}
Qingshan Xu and Wenbing Tao.
\newblock Learning inverse depth regression for multi-view stereo with
  correlation cost volume.
\newblock In {\em Proceedings of the AAAI Conference on Artificial
  Intelligence}, volume~34, pages 12508--12515, 2020.

\bibitem{yan2020dense}
Jianfeng Yan, Zizhuang Wei, Hongwei Yi, Mingyu Ding, Runze Zhang, Yisong Chen,
  Guoping Wang, and Yu-Wing Tai.
\newblock Dense hybrid recurrent multi-view stereo net with dynamic consistency
  checking.
\newblock In {\em European Conference on Computer Vision}, pages 674--689.
  Springer, 2020.

\bibitem{yang2020cost}
Jiayu Yang, Wei Mao, Jose~M Alvarez, and Miaomiao Liu.
\newblock Cost volume pyramid based depth inference for multi-view stereo.
\newblock In {\em Proceedings of the IEEE/CVF Conference on Computer Vision and
  Pattern Recognition}, pages 4877--4886, 2020.

\bibitem{yao2018mvsnet}
Yao Yao, Zixin Luo, Shiwei Li, Tian Fang, and Long Quan.
\newblock Mvsnet: Depth inference for unstructured multi-view stereo.
\newblock In {\em Proceedings of the European Conference on Computer Vision
  (ECCV)}, pages 767--783, 2018.

\bibitem{yao2019recurrent}
Yao Yao, Zixin Luo, Shiwei Li, Tianwei Shen, Tian Fang, and Long Quan.
\newblock Recurrent mvsnet for high-resolution multi-view stereo depth
  inference.
\newblock In {\em Proceedings of the IEEE/CVF Conference on Computer Vision and
  Pattern Recognition}, pages 5525--5534, 2019.

\bibitem{yao2020blendedmvs}
Yao Yao, Zixin Luo, Shiwei Li, Jingyang Zhang, Yufan Ren, Lei Zhou, Tian Fang,
  and Long Quan.
\newblock Blendedmvs: A large-scale dataset for generalized multi-view stereo
  networks.
\newblock In {\em Proceedings of the IEEE/CVF Conference on Computer Vision and
  Pattern Recognition}, pages 1790--1799, 2020.

\bibitem{yi2020PVAMVSNET}
Hongwei Yi, Zizhuang Wei, Mingyu Ding, Runze Zhang, Yisong Chen, Guoping Wang,
  and Yu-Wing Tai.
\newblock Pyramid multi-view stereo net with self-adaptive view aggregation.
\newblock In {\em European Conference on Computer Vision}, pages 766--782.
  Springer, 2020.

\bibitem{zhang2020visibility}
Jingyang Zhang, Yao Yao, Shiwei Li, Zixin Luo, and Tian Fang.
\newblock Visibility-aware multi-view stereo network.
\newblock {\em British Machine Vision Conference (BMVC)}, 2020.

\bibitem{zhu2019deformable}
Xizhou Zhu, Han Hu, Stephen Lin, and Jifeng Dai.
\newblock Deformable convnets v2: More deformable, better results.
\newblock In {\em Proceedings of the IEEE/CVF Conference on Computer Vision and
  Pattern Recognition}, pages 9308--9316, 2019.

\end{thebibliography}
}
\clearpage
\appendix
\appendixpage
\section{Network Details}
Detailed information of the layers of AA-RMVSNet is listed in Tab.~\ref{tab:layers}.  Note that the procedure of feature extraction is identical for all $N$ images and the procedure of cost volume processing is identical for all $D$ depth hypotheses.
\begin{table*}[t]
    \centering
    \begin{tabular}{c|c|c|c}
      \hline
      \hline
      Input  &  Description & Output & Output Shape\\
      \hline
      \hline
      \multicolumn{4}{c}{\textbf{Feature Extraction}: $\mathbf{I}\to \mathbf{f}$} \\
      \multicolumn{4}{c}{$ H\times W\times 3\to  H\times W\times 32$} \\
      \hline
      \hline
      $\mathbf{I}$ & Conv($3\times 3$)+GN+ReLU & $\mathbf{x}_0$ & $ H\times W\times 8$\\
      \hline
      $\mathbf{x}_0$    & Conv($3\times 3$)+GN+ReLU & $\mathbf{x}_1$ & $ H\times W\times 16$\\
      \hline
      $\mathbf{x}_1$    & Conv($3\times 3$)+GN+ReLU & $\mathbf{x}_2$ & $ H\times W\times 16$\\
      \hline
      $\mathbf{x}_2$    & Conv($3\times 3$)+GN+ReLU & $\mathbf{x}_3$ & $ \frac{1}{2}H\times \frac{1}{2}W\times 16$\\
      \hline
      $\mathbf{x}_3$    & Conv($3\times 3$)+GN+ReLU & $\mathbf{x}_4$ & $ \frac{1}{4}H\times \frac{1}{4}W\times 16$\\
      \hline
      $\mathbf{x}_2$ & DeformConv($3\times 3$) & $\mathbf{x}_2'$ & $ H\times W\times 16$ \\
      \hline
      $\mathbf{x}_3$ & DeformConv($3\times 3$)+BI & $\mathbf{x}_3'$ & $ H\times W\times 8$\\
      \hline
      $\mathbf{x}_4$ & DeformConv($3\times 3$)+BI & $\mathbf{x}_4'$ & $ H\times W\times 8$\\
      \hline
      $[\mathbf{x}_2',\mathbf{x}_3',\mathbf{x}_4']$& Concatenation & $\mathbf{f}$ & $H\times W\times 32$ \\
      \hline
      \hline
      \multicolumn{4}{c}{\textbf{Cost Volume Construction}: $\mathbf{f}_{ref,src_{i=1,N-1}}\to\mathbf{C}^{(d)}$} \\
      \multicolumn{4}{c}{$N\times H\times W\times 32 \to H\times W\times 32$} \\
      \hline
      \hline
      $\mathbf{f}_{src_i},d$ & Homography & $\mathbf{f}^{(d)}_{src_i}$ & $ H\times W\times 32$ \\
      \hline
      $\mathbf{f}^{(d)}_{src_i},\mathbf{f}_{ref}$ & $(\mathbf{f}^{(d)}_{src_i}-\mathbf{f}_{ref})^2$ & $\mathbf{c}^{(d)}_i$ & $ H\times W\times 32$ \\
      \hline
      $\mathbf{c}^{(d)}_{i}$   & Conv($3\times 3$)+GN+ReLU   & $\mathbf{x}_5$ & $ H\times W\times 4$  \\
      \hline
      $\mathbf{x}_5$ & Conv($1\times 1$)+GN+ReLU & $\mathbf{x}_6$ & $ H\times W\times 4$ \\
      \hline
      $\mathbf{x}_6$ & Conv($1\times 1$)+GN & $\mathbf{x}_7$ & $ H\times W\times 4$ \\
      \hline
      $\mathbf{x}_5+\mathbf{x}_7$ & ReLU & $\mathbf{x}_8$ & $ H\times W\times 4$ \\
      \hline
      $\mathbf{x}_8$ & Conv($1\times 1$)+Sigmoid & $\boldsymbol{\omega}_i$ & $ H\times W\times 1$ \\
      \hline
      $(1+\boldsymbol{\omega}_i),\mathbf{c}^{(d)}_i$ & $(1+\boldsymbol{\omega}_i)\odot \mathbf{c}^{(d)}_i$ & $\mathbf{c}'^{(d)}_i$ & $ H\times W\times 32$ \\
      \hline
      $\mathbf{c}'^{(d)}_{i=1,\dots,N-1}$ & Arithmetic Mean & $\mathbf{C}^{(d)}$ & $ H\times W\times 32$\\
      \hline
      \hline
      \multicolumn{4}{c}{\textbf{Cost Volume Regularization}: $\mathbf{C}^{(d)}\to \mathbf{Y}^{(d)}$} \\
      \multicolumn{4}{c}{$H\times W\times 32\to H\times W\times 1$} \\
      \hline
      \hline
      $\mathbf{v}^{(d-1)}_0,\mathbf{C}^{(d)}$ & ConvLSTMCell($3\times 3$) & $\mathbf{v}^{(d)}_0$ & $H\times W\times 16$\\
      \hline
      $\mathbf{v}^{(d)}_0$ & MaxPooling & $\mathbf{v}^{(d)}_1$ & $\frac{1}{2}H\times \frac{1}{2}W\times 16$\\
      \hline
      $\mathbf{v}^{(d-1)}_2,\mathbf{v}^{(d)}_1$ & ConvLSTMCell($3\times 3$) & $\mathbf{v}^{(d)}_2$ & $\frac{1}{2}H\times \frac{1}{2}W\times 16$\\
      \hline
      $\mathbf{v}^{(d)}_2$ & MaxPooling & $\mathbf{v}^{(d)}_3$ & $\frac{1}{4}H\times \frac{1}{4}W\times 16$\\
      \hline
      $\mathbf{v}^{(d-1)}_4,\mathbf{v}^{(d)}_3$ & ConvLSTMCell($3\times 3$) & $\mathbf{v}^{(d)}_4$ & $\frac{1}{4}H\times \frac{1}{4}W\times 16$\\
      \hline
      $\mathbf{v}^{(d)}_4$ & TransConv($3\times 3$)+GN+ReLU & $\mathbf{v}^{(d)}_5$ & $\frac{1}{2}H\times \frac{1}{2}W\times 16$\\
      \hline
      $\mathbf{v}^{(d)}_1,\mathbf{v}^{(d)}_5$ & Concatenation & $\mathbf{v}'^{(d)}_5$ & $\frac{1}{2}H\times \frac{1}{2}W\times 32$ \\
      \hline
      $\mathbf{v}^{(d-1)}_6,\mathbf{v}'^{(d)}_5$& ConvLSTMCell($3\times 3$) & $\mathbf{v}^{(d)}_6$ & $\frac{1}{2}H\times \frac{1}{2}W\times 16$\\
      \hline
      $\mathbf{v}^{(d)}_6$ & TransConv($3\times 3$)+GN+ReLU & $\mathbf{v}^{(d)}_7$ & $H\times W\times 16$\\
      \hline
      $\mathbf{v}^{(d)}_0,\mathbf{v}^{(d)}_7$& Concatenation & $\mathbf{v}'^{(d)}_7$ & $H\times W\times 32$ \\
      \hline
      $\mathbf{v}^{(d-1)}_8,\mathbf{v}'^{(d)}_7$ & ConvLSTMCell($3\times 3$) & $\mathbf{v}^{(d)}_8$ & $H\times W\times 8$\\
      \hline
      $\mathbf{v}^{(d)}_8$ & Conv($3\times 3$) & $\mathbf{Y}^{(d)}$ & $H\times W\times 1$\\
      \hline
      \hline
    \end{tabular}
    \caption{Details information of network layers of AA-RMVSNet. Conv, TransConv and DeformConv denote 2D convolution, 2D transposed convolution (also known as deconvolution) and 2D deformable convolution, respectively. GN represents group normalization while BI represents bilinear interpolation.}
    \label{tab:layers}
\end{table*}

\section{Deformable Sampling in Intra-view AA}
In terms of feature extraction for matching, we expect regions with rich texture to be processed by convolutions with smaller receptive fields so that tiny and detailed parts will be preserved during matching. While for low-textured or textureless regions, such as plain surfaces, we prefer a larger receptive field where more context information can get aggregated for more reliable matching. 

\begin{figure}[t]
    \centering
    \includegraphics[width=\columnwidth]{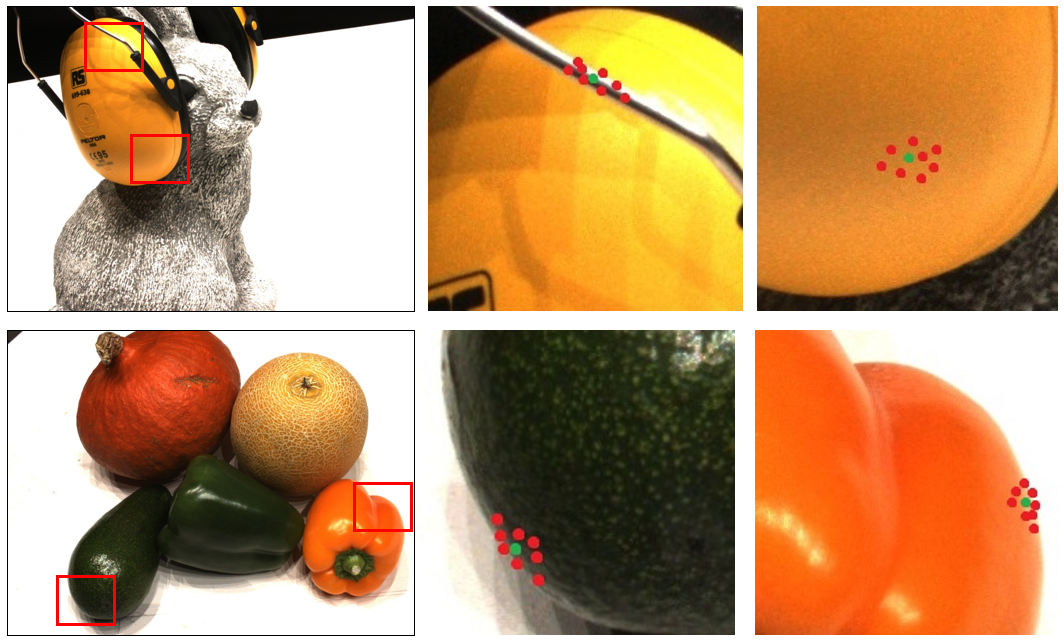}
    \caption{Deformable sampling in different areas, \eg thin object, weak-textured region and object boundary. Green points are centers of convolution kernels and red ones are sampled points with adaptive offsets yielded by sub-networks of deformable convolutions.}
    \label{deform_sampling}
\end{figure}

The proposed intra-view AA module adopts deformable convolution to do the aforementioned job adaptively. For a pixel $\mathbf{p}$ at the object boundary, all sampling points of a deformable convolution kernel tend to be located on the same surface as $\mathbf{p}$. In contrast, for the pixel in textureless regions, sampling points are spread over a larger region and the receptive field is expanded. Fig.~\ref{deform_sampling} visualizes sampling locations of deformable convolution kernels. On the thin cable of the earphone, sampling points tend to be concentrated on the cable itself, while for other low-textured areas of the earphone, the receptive filed is expanded. At boundary regions of objects, sampling points are gathered at the same side of the kernel center.

\section{View Reweighting in Inter-view AA}
In order to handle an arbitrary number of input views and eliminate the influence of unreliable matching at occluded regions, an inter-view AA module is leveraged to our AA-RMVSNet. The inter-view AA module contains a CNN for yielding pixel-wise attention maps for per-view cost volumes adaptively. For an area in the reference image, if this area is occluded in the source image, lower weights should be assigned to suppress local matching. On the contrary, if an area is well-captured and unoccluded, higher weights are assigned to enhance reliable local matching.

\begin{figure}[t]
    \centering
    \includegraphics[width=\columnwidth]{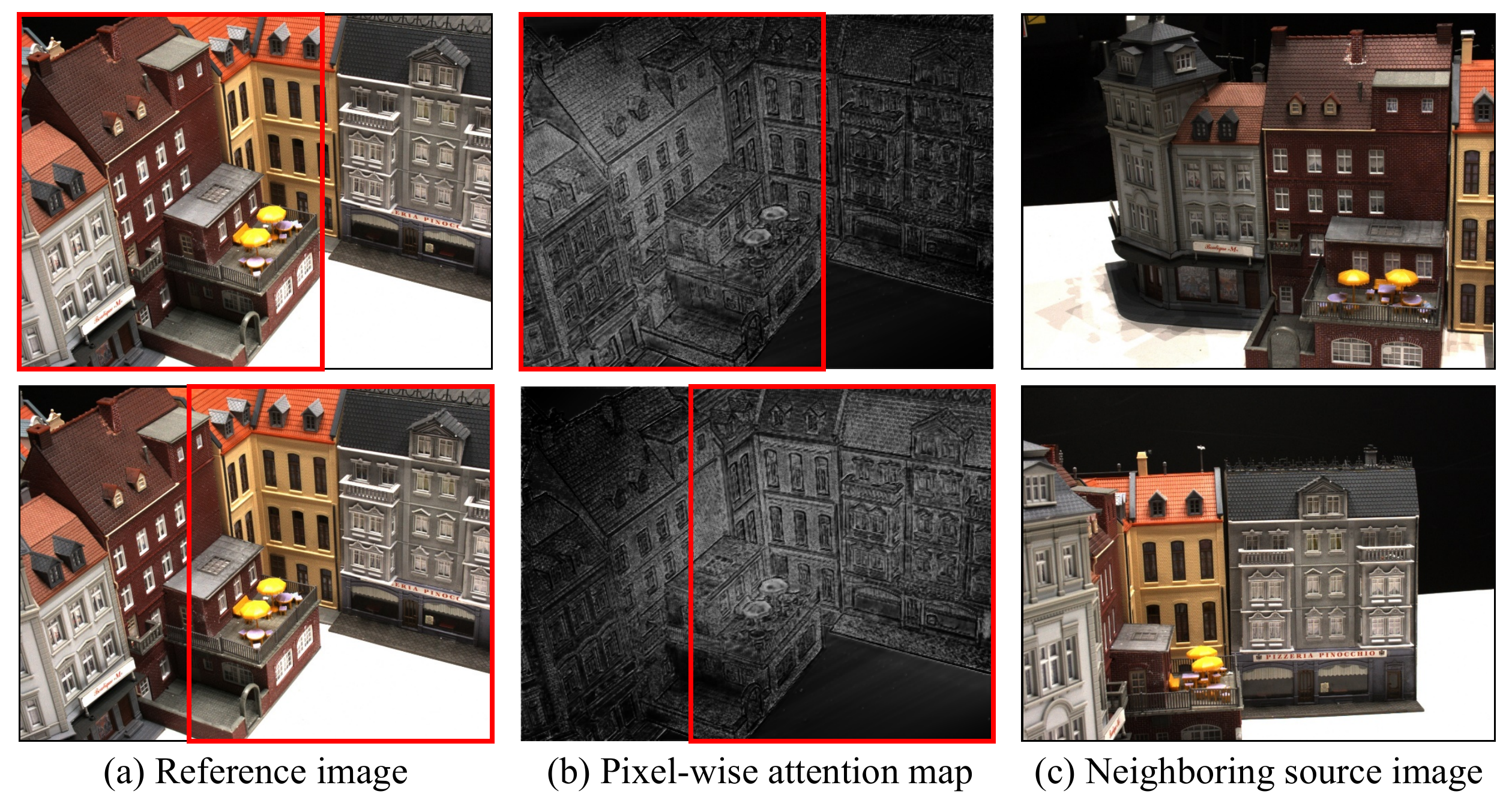}
    \caption{Visualized pixel-wise attention maps yielded by the inter-view AA module. Brighter areas represent higher weights assigned. When matching source images in (c) to the reference image (a), corresponding per-view attention maps are shown as (b).}
    \label{weight}
\end{figure}

Fig.~\ref{weight} visualizes two attention maps by gray-scale images. As is clearly framed in red, for areas well-captured in the corresponding source images, attention values are larger. In this way, reliably matched areas of per-view cost volumes are enhanced while those occluded unreliable regions are suppressed by low weights.


\section{Depth Comparison in Ablation Experiments}
To further demonstrate the effectiveness of the proposed intra-view AA module and inter-view AA module, we visualize some representative depth maps for each ablation experiment.

\begin{figure*}[t]
    \centering
    \includegraphics[width=2\columnwidth]{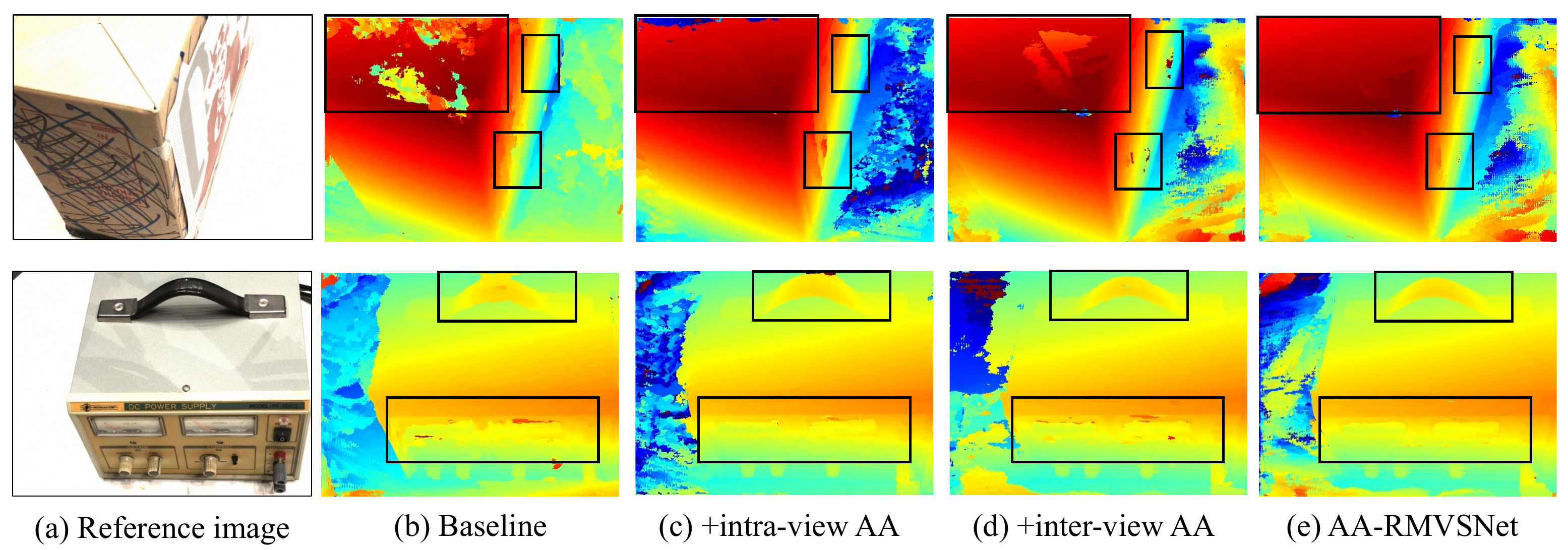}
    \caption{Comparison between depth maps predicted by the network with and without the two proposed AA modules and full AA-RMVSNet.}
    \label{depth_ablation}
\end{figure*}

As is shown in Fig.~\ref{depth_ablation}, the intra-view AA module manages to eliminate noises at textureless surfaces and boundary areas of objects. At the same time, the inter-view AA module is able to preserve more details for those regions easy to be occluded, \eg the handle. Integrated both AA modules into the proposed network, our AA-RMVSNet benefits from both modules and is capable of predicting accurate and complete depth maps for images under varying conditions.

\section{Ablation Study on Experiment Settings}
As is showed in Tab.~\ref{dtu_ablation}, we investigate the influence of variant numbers of input views $N$, numbers of depth hypotheses $D$ and resolutions of input images $W$ and $H$.
\begin{table}[t]
    \footnotesize
    \centering
    \begin{tabular}{ccc|ccc}
     \hline
    $N$ & $D$ & Resolution  & Acc.(mm) & Comp.(mm) & O.A.(mm) \\
    \hline
    3 & 256 & $480\times 360$  & 0.424 & 0.387 & 0.405 \\
    5 & 256 & $480\times 360$  & 0.414 & 0.358 & 0.386 \\
    7 & 256 & $480\times 360$  & 0.408 & 0.351 & 0.380 \\
    7 & 512 & $480\times 360$  & 0.387 & 0.356 & 0.372  \\
    7 & 512 & $640\times 480$  & 0.381 & 0.352 & 0.366 \\
    7 & 512 & $800\times 600$  & \textbf{0.376}& \textbf{0.339} & \textbf{0.357} \\
     \hline
    \end{tabular}
    \caption{Ablation study on number of input views $N$ and number of depth hypotheses $D$ on DTU evaluation set (lower is better).}
    \label{dtu_ablation}
\end{table}

\paragraph{Number of Views}
Our AA-RMVSNet is capable of processing an arbitrary number of views and leveraging the variant importance in multiple views due to the proposed inter-view AA module. With fixed $D$ and image resolution, we compare reconstruction results under $N=3,5,7$. As is shown in Tab.~\ref{dtu_ablation}, the larger $N$ turns, the better the reconstruction results are in terms of all metrics. It demonstrates that our proposed inter-view AA module can well enhance the valid information in the good neighboring views and eliminate bad information in occluded views.
\paragraph{Number of Depth Hypotheses}
In AA-RMVSNet, cost volumes are regularized recurrently by a RNN-CNN hybrid network. In this way, memory usage is reduced considerably and more room is left for finer division of depth space (or known as plane sweep). We compare reconstruction quality when $D=256$ and when $D=512$ with fixed $N=7$ and image resolution $480\times 360$. As a result, finer depth division lowers reconstruction error.
\paragraph{Resolution of Images}
Since our AA-RMVSNet regularizes cost volumes in a memory-efficient fashion, we are able to use images of larger resolution for reconstruction. We fix $N=7$ and $D=512$ and compare reconstruction results with image resolution of $480\times 360$ and $800\times 600$. Experimental results demonstrate that larger resolution is beneficial for reconstruction.

\section{More Point Cloud Results}
\begin{figure*}[t]
    \centering
    \includegraphics[width=\linewidth]{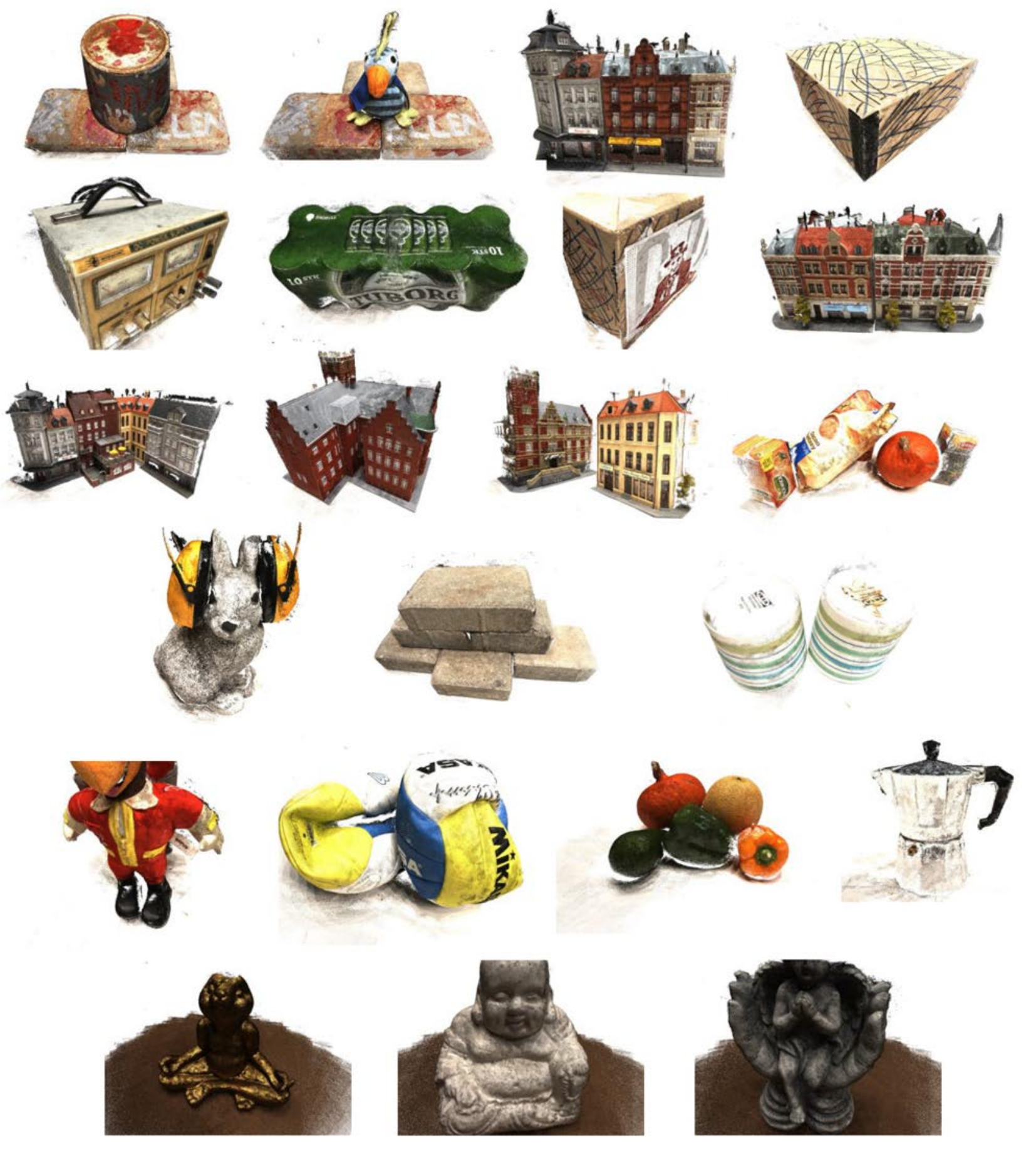}
    \caption{All point clouds results of DTU evaluation set.}
    \label{dtu_all}
\end{figure*}

\begin{figure*}[t]
    \centering
    \includegraphics[width=\linewidth]{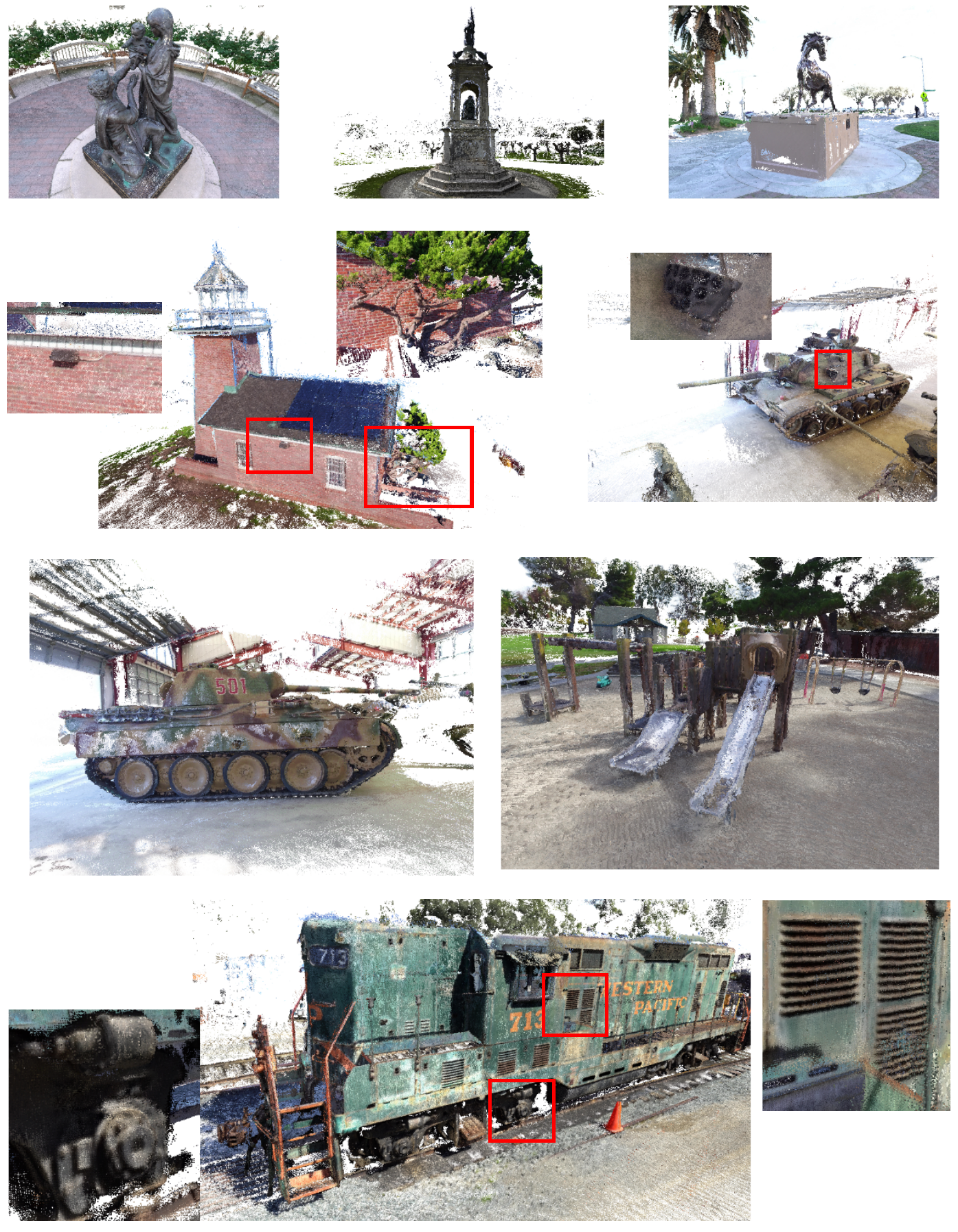}
    \caption{All point clouds results of the intermediate set of Tanks and Temples benchmark.}
    \label{tnt_all}
\end{figure*}

\begin{figure*}[t]
    \centering
    \includegraphics[width=\linewidth]{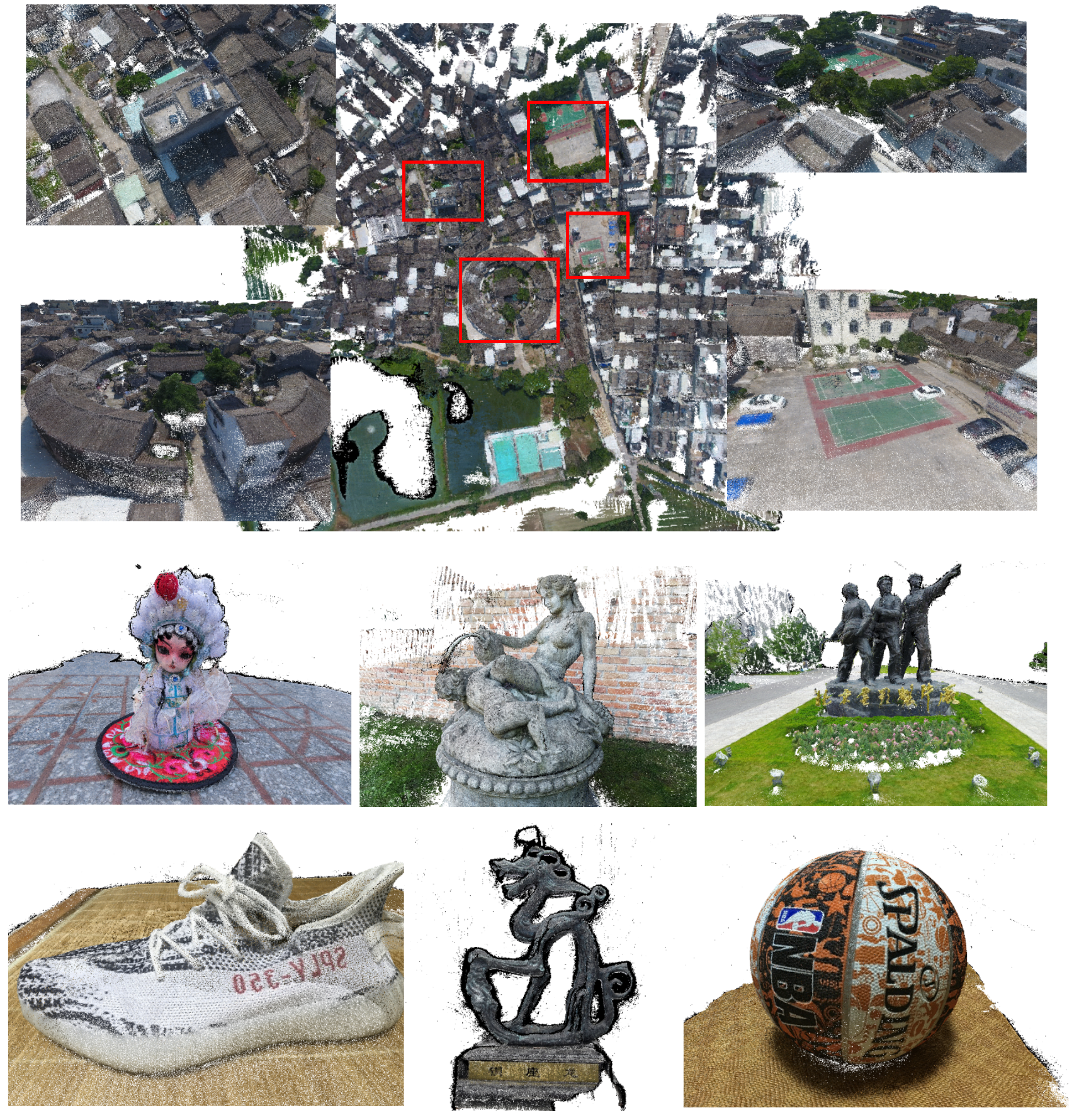}
    \caption{All point clouds results of BlendedMVS validation set.}
    \label{blend_all}
\end{figure*}

We visualize all results of DTU evaluation set, the intermediate set of Tanks and Temples benchmark and BlendedMVS validation set respectively in Fig.~\ref{dtu_all}, Fig.~\ref{tnt_all} and Fig.~\ref{blend_all}. Our AA-RMVSNet demonstrates its robustness and scalability on scenes with varying depth ranges.

\end{document}